\documentclass[11pt]{article}

\usepackage[final]{acl}
\usepackage{times}
\usepackage{latexsym}

\usepackage[T1]{fontenc}

\usepackage[utf8]{inputenc}

\usepackage{microtype}

\usepackage{inconsolata}

\usepackage{graphicx}

\usepackage{algorithm}
\usepackage{enumitem}
\usepackage{newfloat}
\usepackage{listings}
\usepackage[most]{tcolorbox}
\usepackage{xcolor}  
\usepackage{booktabs}
\usepackage{multirow}
\usepackage{amsmath, amssymb}
\usepackage{booktabs}
\usepackage[table]{xcolor}
\usepackage{tabularx}
\usepackage{longtable}
\usepackage{array} 
\usepackage{pifont}
\usepackage{algorithm}
\usepackage{algpseudocode} 

\usepackage{wrapfig}
\usepackage{graphicx} 
\usepackage{xspace}
\usepackage{enumitem}
\usepackage{booktabs}         
\usepackage{multirow}         
\usepackage{colortbl}         
\usepackage[table]{xcolor}    
\usepackage{graphicx}         
\usepackage{algorithm} 
\usepackage{algpseudocode} 
\usepackage{amsmath, amsfonts, amssymb} 
\usepackage{amsthm}

\usepackage{booktabs} 
\usepackage{adjustbox} 
\usepackage{times}
\usepackage{latexsym}
\usepackage{colortbl}
\usepackage{xcolor}
\usepackage{threeparttable}
\usepackage{subcaption}  
\usepackage{threeparttable}
\usepackage{hyperref}
\theoremstyle{remark}

\definecolor{citecolor}{RGB}{2, 56, 189}
\definecolor{myred}{RGB}{207,62,62}
\definecolor{mygreen}{RGB}{112,173,71}
\definecolor{mynewred}{RGB}{178, 56, 40}
\definecolor{rebuttal_blue}{HTML}{1685a9}
\definecolor{rebuttal_purple}{HTML}{9932cd}
\definecolor{rebuttal_red}{HTML}{ef7a82}
\definecolor{avgblue}{RGB}{210,230,250}
\hypersetup{
    colorlinks=true,            
    linkcolor=myred,              
    citecolor=citecolor,            
    urlcolor=magenta            
}
\newcommand{\mname}{\textit{GraphWalker}\xspace} 

\usepackage{booktabs}
\usepackage[table]{xcolor}
\usepackage{colortbl}
\usepackage{multirow}

\newcommand{\pmstd}[2]{%
  #1\ensuremath{\raisebox{0.15ex}{\tiny$\pm$#2}}%
}

\usepackage{xcolor}
\newcommand{\rednum}[1]{\textcolor[HTML]{B22222}{#1}}   
\newcommand{\greennum}[1]{\textcolor[HTML]{228B22}{#1}} 

\newcommand{\newnewcustomfootnotesize}{\fontsize{10pt}{10pt}\selectfont}

\usepackage[table]{xcolor}
\definecolor{ourblue}{RGB}{220,235,250}
\title{\mname{}: Patient Analogy Meets Information Gain for Clinical Reasoning with Large Language Models}

\author{
Yue Fang\textsuperscript{$\spadesuit$} \quad
Weibin Liao\textsuperscript{$\spadesuit$} \quad
Yuxin Guo\textsuperscript{$\spadesuit$} \quad
Jiaran Gao\textsuperscript{$\spadesuit$} \quad
Hongxin Ding\textsuperscript{$\spadesuit$} \\
Jinyang Zhang\textsuperscript{$\spadesuit$} \quad
Xinke Jiang\textsuperscript{$\spadesuit$} \quad
Zhibang Yang\textsuperscript{$\spadesuit$} \\
Junfeng Zhao\textsuperscript{$\spadesuit$}\thanks{Corresponding author.} \quad
Yasha Wang\textsuperscript{$\heartsuit$}\footnotemark[1] \quad
Liantao Ma\textsuperscript{$\heartsuit$}\footnotemark[1] \\
\textsuperscript{$\spadesuit$}School of Computer Science, Peking University, Beijing, China \\
\textsuperscript{$\heartsuit$}National Engineering Research Center for Software Engineering, Peking University, Beijing, China
}

\begin{document}
\maketitle
\begin{abstract}
Clinical reasoning over electronic health records (EHRs) is a 
fundamental yet challenging task in modern healthcare. While 
large language models (LLMs) offer a promising  
paradigm via in-context demonstrations that requires no task-specific parameter updates, existing methods for 
\textbf{reasoning by patient analogy} in EHR settings suffer 
from three core limitations: 
\textbf{(1) Perspective Limitation}, where data-driven 
similarity misaligns with LLM reasoning needs while 
model-driven signals are constrained by limited clinical 
competence; 
\textbf{(2) Cohort Awareness}, as demonstrations are 
selected independently without modeling population-level 
structure; and 
\textbf{(3) Information Aggregation}, where redundancy and 
interaction effects among demonstrations are ignored. 
We propose \mname{}, a training-free framework that lets 
frozen LLMs \textbf{reason by analogy over retrieved patient 
cases}. \mname{} (i) jointly leverages data-driven and 
model-driven perspectives, (ii) discovers patient cohorts to 
ground retrieval in population-level structure, and (iii) 
employs a lazy greedy search with frontier expansion to 
compose demonstrations with high marginal information gain. 
Extensive experiments on multiple real-world EHR benchmarks 
show that \mname{} consistently outperforms state-of-the-art 
demonstration selection baselines, and remains substantially more robust under 
cross-dataset distribution shift, without task-specific parameter updates. \mname{} further generalizes to black-box LLMs and 
composes naturally with agentic reasoning frameworks, 
positioning it as a pluggable \textbf{patient-analogy skill} 
in LLM-based clinical workflows. Our code is available at 
\url{https://github.com/PuppyKnightUniversity/GraphWalker}.
\end{abstract}

\section{Introduction}\label{sec:introduction}
Clinical reasoning on electronic health records (EHRs) 
underpins critical applications such as diagnosis support, 
risk stratification, and treatment planning~\cite{ma2023mortality,
xu2023seqcare,xu2023vecocare,xu2024protomix,fang2023method,
liao2025PAI}. Recent large language models (LLMs) have been 
adapted for EHR clinical reasoning via 
pre-training~\cite{zhang2025adept}, 
post-training~\cite{ding20253ds,fang2025toward,ding2025promed}, 
or hybrid pipelines that distill LLM-derived knowledge into 
supervised EHR models~\cite{jiang2024graphcare,xu2025dearllm}. 
However, these approaches rely on parameter updates over 
task-specific labeled data, with their effectiveness 
constrained by fixed input schemas that are brittle to 
variations in feature order, availability, and encoding 
across heterogeneous EHR systems~\cite{brown2024not,
chen2024clinicalbench,fang2025toward}. Yet clinical practice 
itself relies on a fundamentally different mode of reasoning: 
physicians make decisions by recalling and comparing similar 
past patients, an approach known as \textit{reasoning by 
analogy}~\cite{norman2006building,ten2017principles}. This motivates a lightweight inference-time strategy: equipping a frozen LLM to retrieve and reason over similar past patients, avoiding task-specific parameter updates while remaining robust to distribution shift and composable as a patient-analogy skill in LLM-based clinical workflows.

\begin{figure}[t]
    \centering
    \includegraphics[scale=0.16]{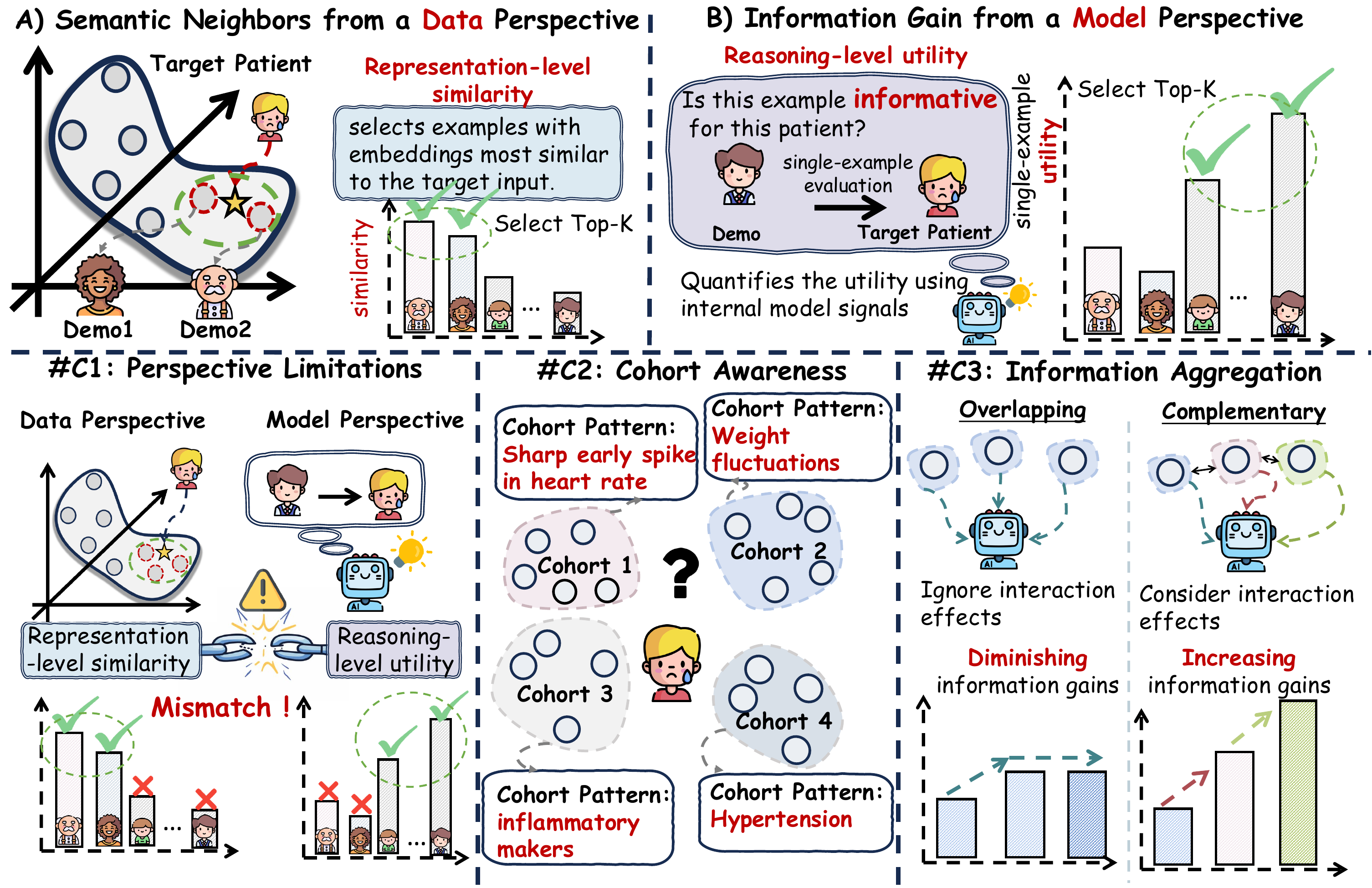}
    \vspace{-0.6cm}\captionsetup{font=footnotesize}
    \caption{Illustration of existing demonstration selection 
    methods (top) and their three core challenges (bottom).}
    \label{fig:introduction}
\vspace{-0.4cm}
\end{figure}

Existing demonstration selection methods fall into two 
perspectives.
\begin{itemize}
[leftmargin=*,itemsep=0pt,parsep=0pt,topsep=0pt,partopsep=0pt]
    \item \textit{\textbf{Semantic Neighbors} from a Data 
    Perspective}: As illustrated in Fig~\ref{fig:introduction}(A), 
    this line~\cite{liu2022makes,hongjin2022selective,
    robertson2009probabilistic} leverages a pre-trained 
    representation model (e.g., SMART)~\cite{yu2024smart} to embed 
    patient records into a latent space, and retrieves the most 
    similar past patients as in-context demonstrations.
    \item \textit{\textbf{Information Gain} from a Model 
    Perspective}: As illustrated in Fig~\ref{fig:introduction}(B), 
    this line~\cite{gonen2023demystifying,peng2024revisiting,
    liumakes,li2025delta} quantifies the utility of candidate 
    demonstrations using LLM-internal signals, such as conditional 
    entropy reduction or gradient-based influence, and selects 
    the top-$k$ demonstrations that maximally improve the model's 
    predictive confidence.
\end{itemize}

Through a systematic analysis of these methods 
(Section~\ref{sec:pilot_study}), we identify \textbf{three core 
challenges} shared by existing works, as shown in 
Fig~\ref{fig:introduction}(C).
\begin{itemize}
[leftmargin=*,itemsep=2pt,parsep=0pt,topsep=2pt,partopsep=0pt]
    \item \textit{Perspective Limitation}: 
    \textit{\textbf{Semantic Neighbors}} implicitly assume that 
    embedding similarity aligns with the LLM's reasoning needs, an 
    assumption that does not consistently hold: highly similar 
    patient trajectories may still contribute little to downstream 
    predictions, suggesting a mismatch between representation-level 
    similarity and reasoning-level utility. 
    \textit{\textbf{Information Gain}}, although more directly 
    aligned with the target model, is inherently constrained by 
    the LLM's clinical competence; when the model lacks sufficient 
    domain knowledge, its internal uncertainty signals may fail to 
    reflect clinically meaningful longitudinal patterns.
    \item \textit{Cohort Awareness}: \textit{\textbf{Both 
    perspectives}} focus on retrieving isolated similar patients, 
    treating each demonstration as independent. However, clinical 
    reasoning is inherently population-driven; selecting 
    individual neighbors without modeling cohort structure can 
    amplify noise from idiosyncratic cases and lead to unstable 
    reasoning contexts.
    \item \textit{Information Aggregation}: \textit{\textbf{Both 
    perspectives}} typically assume that the utility of selected 
    demonstrations accumulates linearly. In practice, multiple 
    similar demonstrations often encode overlapping clinical 
    signals, yielding diminishing marginal gains; ignoring 
    redundancy and interaction effects wastes the limited context 
    window.
\end{itemize}

These challenges motivate a principled framework that bridges 
data and model perspectives, captures cohort structure, and 
accounts for demonstration interactions. To this end, we 
propose \mname{}, a graph-guided framework for 
\textbf{Reasoning by Patient Analogy}, enabling LLMs 
to retrieve and reason over clinically similar patients 
without task-specific parameter updates. \mname{} \textit{integrates data 
and model perspectives} by encoding patient records through 
a pre-trained EHR representation model to construct a 
patient cohort graph, while leveraging LLM-estimated 
information gain to guide traversal and demonstration 
selection. To capture cohort awareness, \mname{} replaces 
instance-level retrieval with \textit{Cohort Discovery}, 
enabling reasoning over clinically coherent patient groups 
rather than isolated cases. To address redundancy and 
interaction among retrieved demonstrations, we introduce 
\textit{Lazy Greedy Search with Frontier Expansion}, 
formulating demonstration selection as a combinatorial 
optimization problem that approximates locally optimal 
demonstration sets under a constrained context budget.

\begin{figure}[!t]
  \centering
\includegraphics[width=0.48\textwidth]{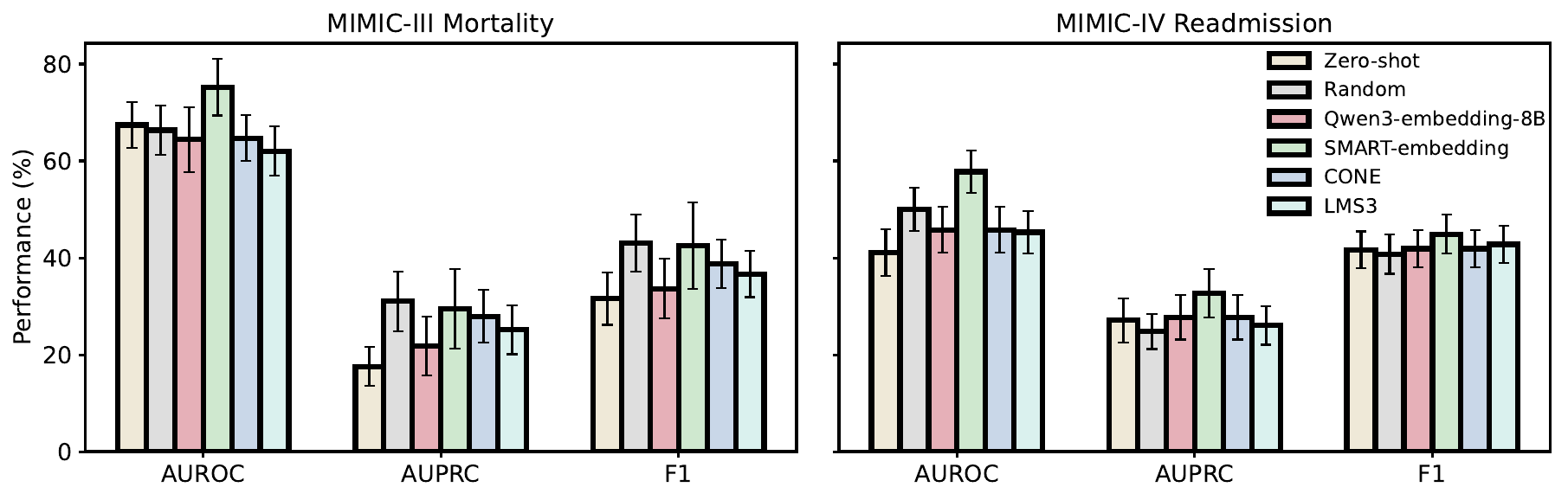}
\vspace{-0.7cm}
  \captionsetup{font=footnotesize}
  \caption{Analysis of the Limitations of a Single Perspective.}
  \label{fig:single_perspective}
  \vspace{-0.4cm}
\end{figure}

\begin{figure}[t]
  \centering
\includegraphics[width=0.48\textwidth]{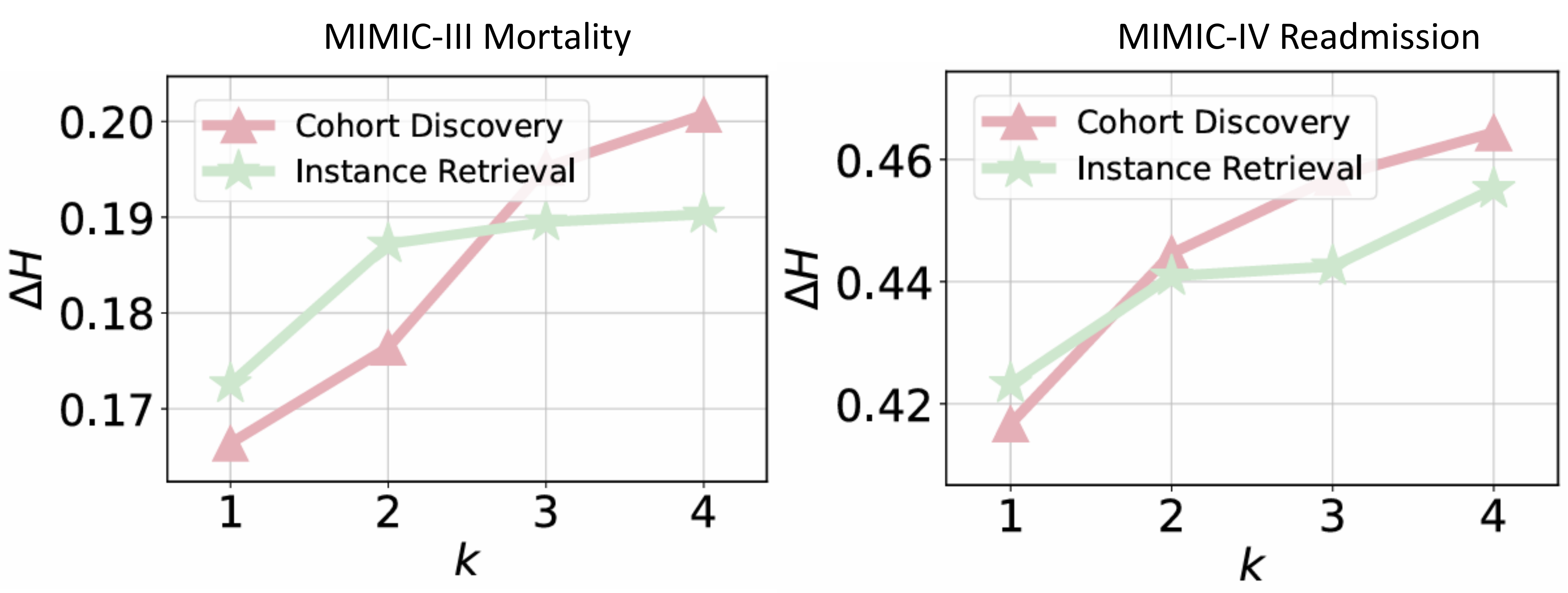}\vspace{-0.4cm}
\captionsetup{font=footnotesize}
  \caption{Analysis of Cohort Discovery \textit{vs.} Instance Retrieval.}
  \label{fig:cohort}
  \vspace{-0.4cm}
\end{figure}

\begin{figure*}[t]
  \centering
\includegraphics[width=0.98\textwidth]{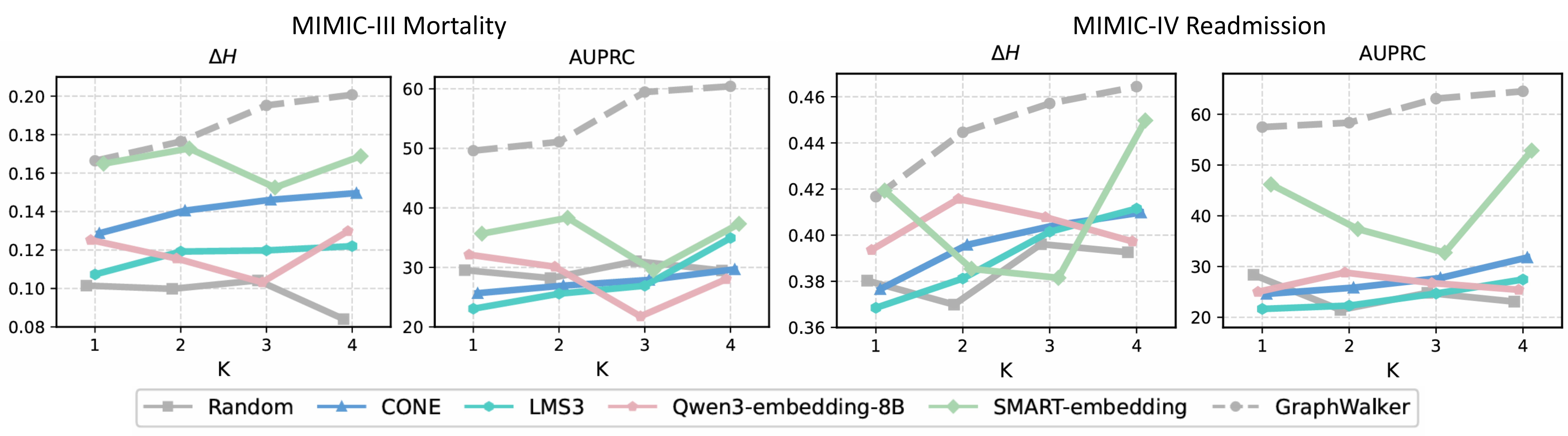}
\vspace{-1.0em}
\captionsetup{font=footnotesize}
  \caption{Analysis of information aggregation across multiple demonstrations under different demonstration selection algorithms.}
  \label{fig:entropy}
  \vspace{-0.4cm}
\end{figure*}

To summarize, our contributions are threefold:
\begin{itemize}
[leftmargin=*,itemsep=2pt,parsep=0pt,topsep=2pt,partopsep=0pt]
    \item \textbf{Insightfully}, through a systematic analysis 
    of existing LLM-based clinical reasoning approaches, we 
    identify three core challenges in demonstration selection 
    for analogical patient reasoning: \textit{Perspective 
    Limitation}, \textit{Cohort Awareness}, and 
    \textit{Information Aggregation}.
    
    \item \textbf{Technically}, we propose \mname{}, a 
    graph-guided framework for reasoning by patient analogy 
    that (1) jointly leverages clinical similarity and 
    LLM-estimated information gain by \textit{integrating 
    data and model perspectives}; (2) employs \textit{Cohort 
    Discovery} to ground retrieval in clinically coherent 
    patient groups rather than isolated cases; and (3) 
    introduces \textit{Lazy Greedy Search with Frontier 
    Expansion} to mitigate diminishing returns in 
    demonstration aggregation.
    
    \item \textbf{Experimentally}, on multiple real-world EHR 
    benchmarks, \mname{} consistently outperforms 
    state-of-the-art demonstration selection baselines, 
    outperforms strong supervised EHR predictors in-domain, and 
    remains substantially more robust under cross-dataset 
    distribution shift, all with only seconds-level inference 
    overhead. \mname{} further generalizes to black-box LLMs 
    and composes naturally with agentic reasoning frameworks, 
    positioning it as a pluggable patient-analogy skill in 
    clinical workflows.
\end{itemize}

\vspace{-0.6em}
\section{Observation \& Motivation}\label{sec:pilot_study}
\vspace{-0.3em}
In this section, we conduct a systematic analysis of 
existing demonstration selection methods and present key 
observations that motivate the design of \mname{}. Detailed 
experimental settings are provided in 
Appendix~\ref{appendix:pilot_setup}.

\vspace{-0.6em}
\subsection{Motivation on Perspective Limitation}
\vspace{-0.3em}

We compare representative methods from both perspectives: 
\textbf{Semantic Neighbors} from a data perspective, instantiated 
as Qwen3-Embedding-8B~\cite{yang2025qwen3} and 
SMART~\cite{yu2024smart} (pre-trained on a large-scale EHR corpus); 
and \textbf{Information Gain} from a model perspective, 
instantiated as CONE~\cite{peng2024revisiting} and 
LMS3~\cite{liumakes}. All methods are evaluated against Zero-shot 
and Random baselines using the same prompt.

As shown in Figure~\ref{fig:single_perspective}, SMART achieves 
the best performance in most cases, confirming the importance of 
clinical knowledge in demonstration selection. Meanwhile, CONE 
and LMS3 consistently outperform the generic 
Qwen3-Embedding-8B, suggesting that information gain serves as 
an effective selection criterion even in the absence of explicit 
clinical knowledge. However, neither perspective dominates: each 
yields strong results on some metrics but unstable performance 
on others. This complementary yet inconsistent behavior 
indicates that no single perspective fully captures the signals 
needed for effective demonstration selection, motivating an 
integrated approach. 

\vspace{-0.6em}
\begin{tcolorbox}[
    enhanced,
    colframe=black!70,
    colback=yellow!5,
    boxrule=0.8pt, arc=2mm,
    left=2mm, right=2mm, top=1.5mm, bottom=1.5mm,
]
\noindent\textbf{Observation I.} 
\textit{Both clinical knowledge (Data Perspective) and information 
gain (Model Perspective) provide valuable but incomplete guidance 
for demonstration selection; effective selection requires 
integrating both perspectives.}
\end{tcolorbox}

\vspace{-1.0em}
\subsection{Motivation on Cohort Awareness}
\vspace{-0.3em}
We compare traditional instance retrieval with our proposed 
\textit{Cohort Discovery} strategy. Using SMART as the EHR 
encoder, we construct a patient graph: instance retrieval 
selects the most similar patient and performs local walks to 
identify remaining demonstrations, whereas \textit{Cohort 
Discovery} first locates the most relevant cohort and then 
selects demonstrations from within it.

As shown in Figure~\ref{fig:cohort}, instance retrieval is 
effective when $k$ is small but becomes less competitive than 
\textit{Cohort Discovery} as $k$ increases. We attribute this to 
the following: a single most-similar patient may be an outlier 
or lie in a sparse cluster, biasing subsequent selection toward 
local optima. In contrast, \textit{Cohort Discovery} leverages 
the relational structure among patients, providing a more robust 
basis for multi-demonstration selection.
\vspace{-0.8em}
\begin{tcolorbox}[
    enhanced,
    colframe=black!70,
    colback=yellow!5,
    boxrule=0.8pt, arc=2mm,
    left=2mm, right=2mm, top=1.5mm, bottom=1.5mm,
]
\noindent\textbf{Observation II.} \textit{Instance retrieval 
overemphasizes a single nearest demonstration during initial 
selection and is prone to local optima; as the number of 
demonstrations increases, its performance degrades relative to 
cohort-based approaches.}
\end{tcolorbox}

\vspace{-1.0em}
\subsection{Motivation on Information Aggregation}
\vspace{-0.3em}
We further analyze how existing methods behave in the 
multi-demonstration setting. As an analysis metric, we 
introduce the LLM's conditional entropy reduction ($\Delta H$, 
see Appendix~\ref{appendix:pilot_setup}), which captures how 
much a given demonstration composition reduces the model's 
predictive uncertainty on the target patient.

Figure~\ref{fig:entropy} reveals two key findings. \textbf{(1)} 
Task performance closely tracks the trend of $\Delta H$, 
indicating that $\Delta H$ serves as an effective guiding 
signal for demonstration selection. This further strengthens 
our motivation to integrate both data-driven and model-driven 
perspectives. \textbf{(2)} \textbf{Semantic Neighbor} methods 
overlook information aggregation: in 
multi-demonstration settings, their performance degrades due 
to information redundancy and conflicts among selected 
demonstrations. \textbf{Information Gain} methods, in contrast, 
maintain monotonic performance improvement as more 
demonstrations are added, yet exhibit pronounced diminishing 
marginal returns due to insufficient consideration of the 
demonstration set as a whole---each demonstration is scored 
independently without modeling interactions with previously 
selected ones.
\vspace{-0.8em}
\begin{tcolorbox}[
    enhanced,
    colframe=black!70,
    colback=yellow!5,
    boxrule=0.8pt, arc=2mm,
    left=2mm, right=2mm, top=1.5mm, bottom=1.5mm,
]
\noindent\textbf{Observation III.} 
\textit{Both \textbf{Semantic Neighbor} and \textbf{Information 
Gain} approaches insufficiently account for Information 
Aggregation, leading to performance degradation or diminishing 
marginal returns in multi-demonstration settings.}
\end{tcolorbox}

\vspace{-1.0em}
\paragraph{Summary.} Based on the above observations, we 
propose \mname, which addresses three key challenges in 
demonstration selection through targeted strategies. 
Specifically, we (1) \textit{integrate data and model 
perspectives} to aggregate clinical knowledge and leverage 
information gain as a guiding signal (for 
\textbf{\textit{Observation I}}), (2) replace 
\textit{Instance Retrieval with Cohort Discovery}, enabling 
reasoning over clinically coherent patient groups rather than 
isolated cases (for \textbf{\textit{Observation II}}), and 
(3) introduce a \textit{Lazy Greedy Search with Frontier 
Expansion}, which formulates demonstration selection as a 
combinatorial optimization problem to address Information 
Aggregation (for \textbf{\textit{Observation III}}).

\begin{figure*}[t]
  \centering
\includegraphics[width=1.02\textwidth]{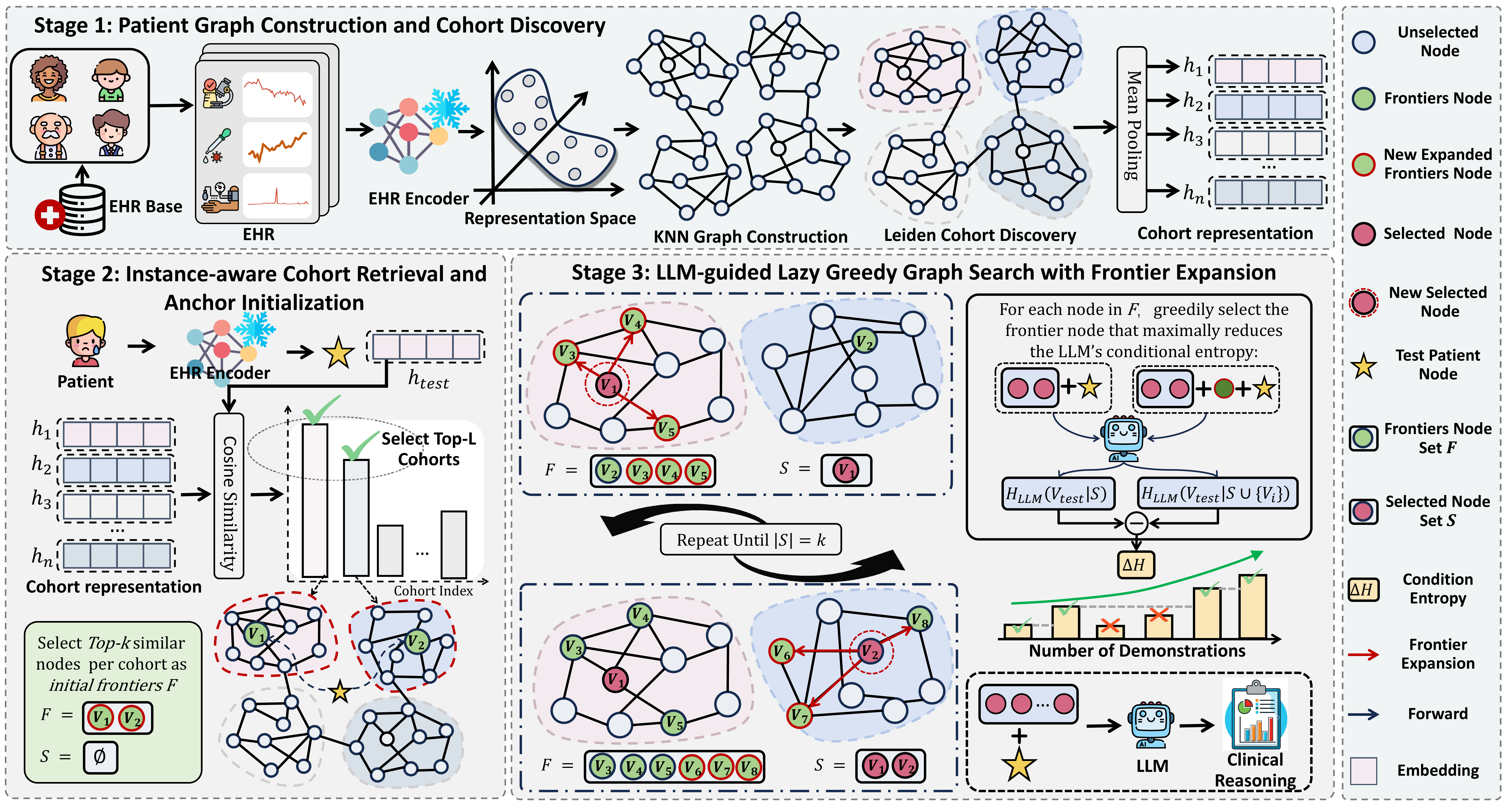}
\vspace{-1.6em}
\captionsetup{font=footnotesize}
  \caption{Illustration of \mname.}
  \label{fig:modelStructure}
  \vspace{-0.4cm}
\end{figure*}

\vspace{-0.6em}
\section{Methodology}
\vspace{-0.3em}
As illustrated in Figure~\ref{fig:modelStructure}, \mname{} 
operates in three stages:
\begin{itemize}[leftmargin=*,noitemsep,topsep=2pt]
    \item \textbf{Patient Graph Construction and Cohort 
    Discovery} builds a population-level patient graph and 
    partitions it into coherent cohorts, capturing shared clinical 
    patterns beyond isolated patients.

    \item \textbf{Instance-aware Cohort Retrieval and 
    Anchor Initialization} retrieves clinically relevant cohorts 
    for the target patient and initializes anchor nodes within 
    them, providing a focused candidate set.

    \item \textbf{LLM-guided Lazy Greedy Search with 
    Frontier Expansion} iteratively constructs the case
    set by selecting frontier nodes that maximize marginal 
    information gain conditioned on the current composition.
\end{itemize}

\vspace{-0.9em}
\subsection{Patient Graph Construction and Cohort Discovery}
\vspace{-0.4em}

\noindent\textbf{Patient Representation.}
To capture clinical patterns that generic semantic embeddings 
overlook (see Section~\ref{sec:pilot_study}), we adopt a 
pretrained Transformer-based EHR encoder (e.g., 
\textit{SMART}~\cite{yu2024smart}). Formally, each patient's 
EHR record is a sequence of $T$ time-ordered visits 
$\mathbf{X} = [\mathbf{x}_1, \ldots, \mathbf{x}_T]$. Given an 
EHR base of $N$ patient records $\{\mathbf{X}_i\}_{i=1}^N$ 
(i.e., the training set), the encoder $\mathcal{M}_{\text{exp}}$ 
maps each record to a fixed-dimensional embedding 
$\mathbf{h}_i = \mathcal{M}_{\text{exp}}(\mathbf{X}_i) \in 
\mathbb{R}^d$, which serves as the patient representation for 
subsequent graph construction.

\noindent\textbf{Population-level Graph Construction.}
We construct a $k$-nearest-neighbor graph $\mathcal{G} = 
(\mathcal{V}, \mathcal{E})$ over patient embeddings to encode 
local clinical similarity at the population level. Each node 
$v_i \in \mathcal{V}$ corresponds to a patient, and an edge 
$(i,j) \in \mathcal{E}$ is established iff patient $j$ is 
among the $k_g$ nearest neighbors of patient $i$ under cosine 
similarity, i.e., $\mathcal{E} = \{ (i,j) \mid \mathrm{rank}_j\, 
\mathrm{sim}(\mathbf{h}_i, \mathbf{h}_j) \le k_g \}$, where 
$\mathrm{sim}(\cdot, \cdot)$ denotes cosine similarity and 
$k_g$ controls graph density.

\noindent\textbf{Graph-based Cohort Discovery.}
To capture shared clinical patterns at the cohort level, we 
partition the patient graph into coherent cohorts of similar 
patients. Specifically, we adopt the \textit{Leiden} 
algorithm~\cite{traag2019louvain} because its well-connected 
community partitions are well suited for constructing stable 
cohort prototypes and supporting subsequent graph-based frontier 
expansion. This produces a set of cohort subgraphs 
$\{\mathcal{C}_m\}_{m=1}^M$ by maximizing graph modularity 
$\mathcal{Q}$:
\begin{equation}
\newnewcustomfootnotesize
\label{eq:modularity}
\mathcal{Q} = \frac{1}{2|\mathcal{E}|} \sum_{i,j} \left( A_{ij} 
- \frac{d_i d_j}{2|\mathcal{E}|} \right) \mathbb{I}(c_i = c_j),
\end{equation}
where $A_{ij}$ is the adjacency matrix of $\mathcal{G}$, $d_i$ is 
the degree of node $i$, and $\mathbb{I}(\cdot)$ indicates whether 
two nodes belong to the same cohort. Each cohort exhibits dense 
internal connectivity and sparse external connectivity. We 
summarize each cohort $\mathcal{C}_m$ by mean-pooling its member 
embeddings into a cohort prototype 
$\mathbf{z}_m = \frac{1}{|\mathcal{C}_m|} \sum_{i \in \mathcal{C}_m} \mathbf{h}_i$, 
which serves as a compact cohort representation for 
instance-aware cohort retrieval.

\vspace{-0.3em}
\subsection{Instance-aware Cohort Retrieval and Anchor Initialization}
\vspace{-0.3em}
\noindent\textbf{Instance-aware Cohort Retrieval.}
Given a target patient with EHR record $\mathbf{X}^{(q)}$, we 
obtain its embedding $\mathbf{h}^{(q)} = 
\mathcal{M}_{\text{exp}}(\mathbf{X}^{(q)})$ using the same 
pretrained EHR encoder as in Stage~1. To account for cohort 
awareness and avoid reasoning over isolated patient instances, we 
retrieve clinically relevant cohorts by aligning $\mathbf{h}^{(q)}$ 
with cohort prototypes, thereby grounding the subsequent 
case search in population-level shared clinical patterns. 
Specifically, we select the top-$K_c$ cohorts whose prototypes 
$\mathbf{z}_m$ are most similar to $\mathbf{h}^{(q)}$:
\begin{equation}
\newnewcustomfootnotesize
\label{eq:cohort_retrieval}
\mathcal{C}^{(q)}_{\text{ret}} = \operatorname{Top}\text{-}K_c 
\bigl( \mathrm{sim}(\mathbf{h}^{(q)}, \mathbf{z}_m) \bigr).
\end{equation}
This step situates the target patient within clinically coherent 
regions of the patient graph, providing a focused context for 
the subsequent step.

\noindent\textbf{Anchor Initialization.}
Given the retrieved cohorts $\mathcal{C}^{(q)}_{\text{ret}}$, we 
initialize the search frontier by selecting anchor nodes within 
each cohort. Specifically, for each cohort $\mathcal{C}_m \in 
\mathcal{C}^{(q)}_{\text{ret}}$, we select the top-$K_a$ patients 
whose embeddings are most similar to $\mathbf{h}^{(q)}$:
\begin{equation}
\newnewcustomfootnotesize
\label{eq:anchor_init}
\mathcal{V}^{(q)}_m = \operatorname{Top}\text{-}K_a 
\bigl( \mathrm{sim}(\mathbf{h}^{(q)}, \mathbf{h}_i) 
\bigr)_{v_i \in \mathcal{C}_m}.
\end{equation}
The initial search frontier is then defined as the union of 
anchors across retrieved cohorts: 
$\mathcal{F}^{(q)} = \bigcup_{\mathcal{C}_m \in 
\mathcal{C}^{(q)}_{\text{ret}}} \mathcal{V}^{(q)}_m$.

\vspace{-0.3em}
\subsection{LLM-guided Lazy Greedy Search with Frontier Expansion}
\vspace{-0.3em}

\noindent\textbf{LLM-guided Greedy Selection.}
Starting from the initial frontier $\mathcal{F}^{(q)}$, we 
iteratively expand the demonstration set $\mathcal{S}^{(q)}$ 
by selecting candidates that maximize the marginal 
\emph{information gain} for the target patient. We extend 
the entropy-based information gain to a 
\textit{composition-conditioned} criterion that scores each 
candidate against the running demonstration set rather than 
independently, an aspect overlooked by prior work. 

Formally, let $H_\theta(x_{\text{test}} \mid \mathcal{S})$ denote 
the conditional entropy of the LLM's prediction on the target 
query $x_{\text{test}}$ given a demonstration composition 
$\mathcal{S}$, capturing the model's predictive uncertainty under 
that composition. For a candidate node $v_i \in \mathcal{F}^{(q)}$, 
its \emph{marginal information gain} with respect to the current 
demonstration set $\mathcal{S}^{(q)}$ is defined as
\begin{equation}
\newnewcustomfootnotesize
\label{eq:marginal_gain}
\Delta H_\theta(v_i \mid \mathcal{S}^{(q)}) = 
H_\theta(x_{\text{test}} \mid \mathcal{S}^{(q)}) - 
H_\theta(x_{\text{test}} \mid \mathcal{S}^{(q)} \cup \{v_i\}).
\end{equation}
A larger $\Delta H_\theta(\cdot)$ indicates that $v_i$ provides 
more complementary information that enhances the collective 
reasoning utility of the current composition. At each step, we 
select the frontier node with the largest marginal gain and add 
it to the demonstration set:
\begin{equation}
\newnewcustomfootnotesize
\label{eq:greedy_select_update}
v^\ast = \arg\max_{v_i \in \mathcal{F}^{(q)}} 
\Delta H_\theta \bigl( v_i \mid \mathcal{S}^{(q)} \bigr), 
\quad \mathcal{S}^{(q)} \leftarrow \mathcal{S}^{(q)} \cup 
\{v^\ast\}.
\end{equation}
To avoid recomputing $\Delta H_\theta$ for every frontier node at 
each iteration, we implement this procedure with a \emph{lazy 
greedy} strategy, maintaining a priority queue with cached 
marginal gains and recomputing them only when 
necessary~\citep{minoux2005accelerated}; further details are 
provided in Appendix~\ref{app:lazy_frontier_greedy}. The updated 
demonstration set then triggers the graph-based frontier 
expansion step.

\noindent\textbf{Graph-based Frontier Expansion.}
After each selection step, we expand the search frontier to 
introduce new candidates that are conditionally relevant to the 
current demonstration composition. Specifically, given the newly 
selected node $v^\ast$, we add its graph neighbors to the 
frontier while excluding previously selected nodes:
\begin{equation}
\newnewcustomfootnotesize
\label{eq:frontier_expand}
\mathcal{F}^{(q)} \leftarrow \bigl(\mathcal{F}^{(q)} \cup 
\mathcal{N}(v^\ast)\bigr) \setminus \mathcal{S}^{(q)},
\end{equation}
where $\mathcal{N}(v^\ast) = \{ v_j \mid (v^\ast, v_j) \in 
\mathcal{E} \}$ denotes the neighbors of $v^\ast$ in the patient 
graph. This expansion exploits the local connectivity of the 
patient graph, where neighboring nodes are expected to share 
locally coherent clinical patterns. Conditioned on the fact that 
$v^\ast$ yields high information gain for the target query, its 
neighbors are more likely to provide complementary clinical 
evidence that further enhances the reasoning utility of the 
current demonstration composition.

\noindent\textbf{Stopping Criterion.}
The procedure terminates when either (i) a 
predefined budget of $K$ demonstrations is reached, or (ii) no 
candidate in the current frontier yields a positive marginal 
information gain, preventing the inclusion of misleading 
or uninformative demonstrations.

\noindent\textbf{LLM Inference with Selected Demonstrations.}
After selection, the demonstration set $\mathcal{S}^{(q)}$ is 
used to construct a few-shot prompt (see 
Appendix~\ref{app:prompt} for details), which is then fed to the 
target LLM to generate the final prediction. The full algorithm 
is provided in Appendix~\ref{app:algorithm}.

\begin{table*}[!ht]
\centering
\fontsize{7.5pt}{8pt}\selectfont
\setlength{\tabcolsep}{4.0pt}
\renewcommand{\arraystretch}{1.1}

\begin{tabular}{l|l|ccc|cc|ccc}
\toprule
\multicolumn{2}{c|}{\textbf{Method}}
& \multicolumn{3}{c|}{\textbf{MIMIC-III Mortality}}
& \multicolumn{2}{c|}{\textbf{MIMIC-III LOS}}
& \multicolumn{3}{c}{\textbf{MIMIC-IV Readmission}} \\
\textbf{Paradigm} & \textbf{Approach}
& AUROC $\uparrow$ & AUPRC $\uparrow$ & F1 $\uparrow$
& ma-ROC $\uparrow$ & mi-ROC $\uparrow$
& AUROC $\uparrow$ & AUPRC $\uparrow$ & F1 $\uparrow$ \\
\midrule

\rowcolor{gray!10}
\multicolumn{10}{c}{\textit{Qwen3-14B}} \\
\midrule
\multirow{2}{*}{Vanilla} & Zero-shot
& \pmstd{64.09}{5.26} & \pmstd{17.58}{3.98} & \pmstd{31.62}{5.44}
& \pmstd{65.22}{3.00} & \pmstd{37.49}{2.48}
& \pmstd{41.11}{4.77} & \pmstd{27.11}{4.51} & \pmstd{41.66}{3.80} \\
& Random
& \pmstd{66.33}{5.12} & \pmstd{31.04}{6.89} & \pmstd{43.00}{5.88}
& \pmstd{58.12}{1.98} & \pmstd{61.28}{3.35}
& \pmstd{49.98}{4.47} & \pmstd{24.81}{3.63} & \pmstd{40.77}{4.05} \\
\midrule
\multirow{3}{*}{Data Perspective} & Semantic-emb
& \pmstd{64.43}{6.71} & \pmstd{21.78}{6.09} & \pmstd{33.60}{6.16}
& \pmstd{57.06}{3.77} & \pmstd{66.60}{2.60}
& \pmstd{45.12}{4.13} & \pmstd{26.91}{4.99} & \pmstd{42.01}{3.94} \\
& EHR-emb
& \pmstd{75.24}{5.86} & \pmstd{29.52}{8.20} & \pmstd{42.49}{8.88}
& \pmstd{62.03}{3.13} & \pmstd{66.98}{2.53}
& \underline{\pmstd{57.79}{4.37}} & \underline{\pmstd{32.73}{5.01}} & \underline{\pmstd{44.87}{4.01}} \\
& Time-series
& \pmstd{62.28}{5.09} & \pmstd{29.62}{6.10} & \pmstd{39.34}{6.13}
& \pmstd{58.03}{3.50} & \pmstd{55.15}{2.62}
& \pmstd{53.46}{4.70} & \pmstd{26.97}{4.05} & \pmstd{40.55}{3.91} \\
\midrule
\multirow{7}{*}{Model Perspective} & SPELL
& \underline{\pmstd{75.37}{4.83}} & \underline{\pmstd{32.23}{6.55}} & \underline{\pmstd{50.32}{6.86}}
& \pmstd{63.73}{3.65} & \underline{\pmstd{75.78}{2.18}}
& \pmstd{48.82}{4.37} & \pmstd{24.37}{3.38} & \pmstd{40.03}{3.98} \\
& Influence
& \pmstd{60.54}{5.64} & \pmstd{27.20}{6.59} & \pmstd{36.10}{6.88}
& \pmstd{50.03}{2.64} & \pmstd{62.70}{2.30}
& \pmstd{52.85}{3.28} & \pmstd{28.55}{3.75} & \pmstd{43.19}{3.85} \\
& IDS
& \pmstd{59.14}{4.53} & \pmstd{19.73}{3.93} & \pmstd{32.97}{4.45}
& \pmstd{49.86}{2.06} & \pmstd{37.41}{1.11}
& \pmstd{47.41}{4.51} & \pmstd{25.43}{3.48} & \pmstd{42.68}{3.79} \\
& CONE
& \pmstd{64.69}{4.74} & \pmstd{27.90}{5.44} & \pmstd{38.82}{5.01}
& \pmstd{57.84}{3.66} & \pmstd{67.91}{2.57}
& \pmstd{45.76}{4.72} & \pmstd{27.75}{4.60} & \pmstd{41.89}{3.79} \\
& LMS3
& \pmstd{66.54}{5.61} & \pmstd{26.92}{6.55} & \pmstd{39.07}{7.46}
& \pmstd{59.44}{3.21} & \pmstd{31.89}{2.54}
& \pmstd{44.34}{4.57} & \pmstd{24.68}{3.47} & \pmstd{42.28}{3.81} \\
& Delta-KNN
& \pmstd{51.52}{6.34} & \pmstd{15.66}{4.31} & \pmstd{25.22}{4.87}
& \pmstd{59.04}{3.61} & \pmstd{63.43}{2.51}
& \pmstd{47.11}{4.81} & \pmstd{24.41}{3.90} & \pmstd{40.00}{3.91} \\
& GradSel
& \pmstd{62.09}{5.40} & \pmstd{32.20}{7.03} & \pmstd{36.68}{5.68}
& \underline{\pmstd{66.80}{3.37}} & \pmstd{73.57}{2.01}
& \pmstd{45.73}{4.29} & \pmstd{25.39}{3.44} & \pmstd{41.83}{3.81} \\
\midrule
\rowcolor[HTML]{F0F6FF}
Ours & \textit{GraphWalker}
& \textbf{\pmstd{84.19}{3.35}} & \textbf{\pmstd{59.46}{7.64}} & \textbf{\pmstd{58.24}{5.89}}
& \textbf{\pmstd{69.51}{3.16}} & \textbf{\pmstd{84.64}{2.14}}
& \textbf{\pmstd{75.91}{4.11}} & \textbf{\pmstd{60.61}{6.19}} & \textbf{\pmstd{58.01}{5.33}} \\
\midrule

\rowcolor{gray!10}
\multicolumn{10}{c}{\textit{LLaMA-3.1-8B-Instruct}} \\
\midrule
\multirow{2}{*}{Vanilla} & Zero-shot
& \pmstd{65.35}{4.90} & \pmstd{21.72}{4.61} & \pmstd{35.75}{5.22}
& \pmstd{43.16}{4.25} & \pmstd{74.13}{2.63}
& \pmstd{54.40}{5.02} & \pmstd{29.88}{4.99} & \pmstd{41.24}{4.14} \\
& Random
& \pmstd{52.04}{5.55} & \pmstd{18.81}{4.78} & \pmstd{28.90}{4.11}
& \pmstd{45.66}{4.24} & \pmstd{80.14}{2.31}
& \pmstd{48.63}{5.03} & \pmstd{27.06}{4.68} & \pmstd{39.85}{3.85} \\
\midrule
\multirow{3}{*}{Data Perspective} & Semantic-emb
& \pmstd{56.90}{5.44} & \pmstd{22.32}{4.69} & \pmstd{35.15}{5.10}
& \pmstd{53.78}{3.87} & \pmstd{81.49}{2.26}
& \pmstd{43.55}{4.33} & \pmstd{23.69}{3.25} & \pmstd{42.37}{3.83} \\
& EHR-emb
& \underline{\pmstd{65.78}{4.24}} & \pmstd{24.30}{4.39} & \underline{\pmstd{41.89}{5.14}}
& \pmstd{45.72}{4.08} & \pmstd{74.78}{2.63}
& \underline{\pmstd{63.75}{4.78}} & \underline{\pmstd{46.32}{6.37}} & \underline{\pmstd{49.81}{4.88}} \\
& Time-series
& \pmstd{56.38}{6.06} & \pmstd{24.86}{5.56} & \pmstd{37.16}{5.78}
& \pmstd{46.23}{3.73} & \pmstd{71.99}{2.57}
& \pmstd{53.45}{4.51} & \pmstd{27.67}{4.43} & \pmstd{42.28}{4.28} \\
\midrule
\multirow{7}{*}{Model Perspective} & SPELL
& \pmstd{55.43}{4.83} & \pmstd{20.92}{4.44} & \pmstd{34.51}{4.45}
& \pmstd{49.48}{4.07} & \pmstd{82.02}{2.24}
& \pmstd{48.11}{4.67} & \pmstd{25.03}{4.09} & \pmstd{40.09}{3.86} \\
& Influence
& \pmstd{61.73}{5.31} & \underline{\pmstd{31.80}{6.55}} & \pmstd{38.39}{5.19}
& \underline{\pmstd{55.57}{4.39}} & \underline{\pmstd{82.63}{2.37}}
& \pmstd{49.64}{3.99} & \pmstd{28.40}{4.17} & \pmstd{42.65}{3.81} \\
& IDS
& \pmstd{59.89}{5.52} & \pmstd{28.40}{6.61} & \pmstd{36.11}{5.13}
& \pmstd{42.12}{3.58} & \pmstd{71.14}{2.10}
& \pmstd{45.25}{4.52} & \pmstd{24.86}{3.59} & \pmstd{42.65}{3.81} \\
& CONE
& \pmstd{50.40}{5.29} & \pmstd{18.93}{3.97} & \pmstd{32.79}{4.54}
& \pmstd{38.31}{3.82} & \pmstd{76.94}{2.53}
& \pmstd{51.67}{4.73} & \pmstd{29.42}{4.45} & \pmstd{42.73}{3.86} \\
& LMS3
& \pmstd{51.80}{5.73} & \pmstd{16.41}{3.38} & \pmstd{29.37}{4.67}
& \pmstd{39.60}{3.86} & \pmstd{74.05}{2.68}
& \pmstd{54.77}{4.81} & \pmstd{30.95}{4.55} & \pmstd{45.41}{4.73} \\
& Delta-KNN
& \pmstd{57.42}{4.91} & \pmstd{21.48}{4.12} & \pmstd{33.61}{4.43}
& \pmstd{47.69}{3.82} & \pmstd{75.71}{2.60}
& \pmstd{51.23}{4.62} & \pmstd{26.50}{4.15} & \pmstd{41.13}{4.00} \\
& GradSel
& \pmstd{47.26}{5.37} & \pmstd{17.32}{3.22} & \pmstd{31.23}{4.04}
& \pmstd{53.92}{4.14} & \pmstd{80.83}{2.29}
& \pmstd{52.33}{4.04} & \pmstd{28.70}{4.02} & \pmstd{42.17}{3.83} \\
\midrule
\rowcolor[HTML]{F0F6FF}
Ours & \textit{GraphWalker}
& \textbf{\pmstd{76.71}{3.58}} & \textbf{\pmstd{35.09}{6.40}} & \textbf{\pmstd{50.13}{5.52}}
& \textbf{\pmstd{57.06}{3.98}} & \textbf{\pmstd{83.55}{2.16}}
& \textbf{\pmstd{68.32}{4.19}} & \textbf{\pmstd{46.37}{6.65}} & \textbf{\pmstd{51.54}{4.45}} \\
\bottomrule
\end{tabular}
\vspace{-0.2cm}
\captionsetup{font=footnotesize}
\caption{Performance comparison on MIMIC-III and MIMIC-IV 
datasets using \textit{Qwen3-14B} and 
\textit{Llama3.1-8B-Instruct} as backbone LLMs. All metrics 
are higher-is-better ($\uparrow$); values are 
mean$\pm$std over three random seeds. \textbf{Bold}: best; 
\underline{underlined}: second-best.}
\label{tab:main_results_mimic}
\vspace{-0.4cm}
\end{table*}
\vspace{-0.6em}
\section{Experiment}
\vspace{-0.3em}
In this section, we conduct extensive experiments to answer: 
(\textbf{RQ1}) Does \mname{} outperform state-of-the-art 
demonstration selection baselines for clinical EHR reasoning? 
(\textbf{RQ2}) How sensitive is \mname{} to its key components 
and hyper-parameters, and what is its runtime efficiency? 
(\textbf{RQ3}) How does parameter-free \mname{} compare against 
supervised EHR predictive models, both in-domain and under 
cross-dataset distribution shift? 
(\textbf{RQ4}) Can \mname{} serve as a composable 
patient-analogy skill within LLM-based agentic clinical 
workflows?

\vspace{-0.3em}
\subsection{Experimental Setup}\label{sec:exp-setup}
\vspace{-0.3em}
\paragraph{Datasets and Tasks.} 
We evaluate \mname{} on two widely-used EHR benchmarks: 
\textbf{MIMIC-III}~\cite{johnson2016mimic} and 
\textbf{MIMIC-IV}~\cite{johnson2023mimic}, across three standard 
clinical prediction tasks: in-hospital mortality, length-of-stay 
(LOS), and ICU readmission. Dataset statistics, task definitions, 
and preprocessing details are in Appendix~\ref{app:dataset} 
and~\ref{app:task}.

\noindent\textbf{Baselines.} 
We compare \mname{} against representative demonstration 
selection baselines from two perspectives. 
(1) Data perspective: 
\textit{Semantic-emb}~\cite{zhang2025qwen3}, 
\textit{EHR-emb}~\cite{yu2024smart}, and 
\textit{Time-series}~\cite{kawarada2024demonstration}. 
(2) Model perspective: 
\textit{SPELL}~\cite{gonen2023demystifying}, 
\textit{Influence}~\cite{nguyen2023context}, 
\textit{IDS}~\cite{qin2024context}, 
\textit{CONE}~\cite{peng2024revisiting}, 
\textit{LMS3}~\cite{liumakes}, 
\textit{GradSel}~\cite{zhang2025linear}, and 
\textit{Delta-KNN}~\cite{li2025delta}. 
We also include \textit{Zero-shot} and \textit{Random} as 
reference baselines. Implementation details of baselines and ours are in 
Appendix~\ref{appen:baseline_details} and ~\ref{appendix:Implementation}.

\noindent\textbf{Metrics.} 
We report AUROC, AUPRC, and F1-score for mortality and 
readmission; following~\citet{harutyunyan2019multitask}, we 
report macro- and micro-averaged ROC for LOS. 

\vspace{-0.3em}
\subsection{Main Results (RQ1)}
\label{sec:main_results}
\vspace{-0.3em}
We report the main results with \textit{Qwen3-14B} and 
\textit{LLaMA-3.1-8B-Instruct} in 
Table~\ref{tab:main_results_mimic}, and defer the full results 
with the larger \textit{Qwen3-32B} backbone to Appendix 
Table~\ref{tab:appendix_qwen32b_results}. We summarize the key 
observations as follows:

\noindent\textbf{Takeaway \ding{202}: \mname{} consistently 
achieves the strongest performance across backbones and 
tasks.} Over the second-best baseline, \mname{} attains an 
average absolute improvement of \rednum{+9.93$\uparrow$} in 
AUROC, \rednum{+12.99$\uparrow$} in AUPRC, 
\rednum{+7.89$\uparrow$} in F1, \rednum{+2.86$\uparrow$} in 
ma-ROC, and \rednum{+4.31$\uparrow$} in mi-ROC.

\noindent\textbf{Takeaway \ding{203}: Dual-perspective 
selection outperforms all single-perspective baselines.} 
\mname{} consistently outperforms existing 
\textit{single-perspective} methods, empirically validating 
the advantage of a dual-view demonstration selection 
paradigm that bridges clinical similarity and LLM-internal 
reasoning dynamics.

\vspace{-0.3em}
\subsection{Framework and Efficiency Analysis (RQ2)}
\vspace{-0.3em}

\begin{table*}[t!]
\centering
\fontsize{7.5pt}{8pt}\selectfont
\setlength{\tabcolsep}{4.0pt}
\renewcommand{\arraystretch}{1.1}

\begin{tabular}{l|ccc|cc|ccc}
\toprule
\textbf{Method }
& \multicolumn{3}{c|}{\textbf{MIMIC-III Mortality}} 
& \multicolumn{2}{c|}{\textbf{MIMIC-III LOS}} 
& \multicolumn{3}{c}{\textbf{MIMIC-IV Readmission}} \\
& AUROC $\uparrow$ & AUPRC $\uparrow$ & F1 $\uparrow$
& ma-ROC $\uparrow$ & mi-ROC $\uparrow$
& AUROC $\uparrow$ & AUPRC $\uparrow$ & F1 $\uparrow$ \\
\midrule

\rowcolor[HTML]{F0F6FF}
\mname
& \textbf{\pmstd{84.19}{3.35}} & \textbf{\pmstd{59.46}{7.64}} & \textbf{\pmstd{58.24}{5.89}}
& \textbf{\pmstd{69.51}{3.16}} & \textbf{\pmstd{84.64}{2.14}}
& \textbf{\pmstd{75.91}{4.11}} & \textbf{\pmstd{60.61}{6.19}} & \textbf{\pmstd{58.01}{5.33}} \\
\midrule

\textit{w/o EHR-emb}
& \pmstd{61.09}{5.32} & \pmstd{31.13}{6.81} & \pmstd{37.50}{5.73}
& \pmstd{62.03}{3.40} & \pmstd{72.63}{2.46}
& \pmstd{45.30}{4.55} & \pmstd{26.69}{4.20} & \pmstd{41.74}{3.79} \\

\textit{w/o Cohort}
& \pmstd{81.19}{3.78} & \pmstd{48.27}{8.01} & \pmstd{55.06}{6.18}
& \pmstd{65.85}{3.75} & \pmstd{81.16}{2.27}
& \pmstd{75.65}{4.04} & \pmstd{59.61}{6.09} & \pmstd{56.87}{5.19} \\

\textit{w/o Greedy \& Frontier Expansion}
& \pmstd{69.43}{5.80} & \pmstd{46.72}{8.24} & \pmstd{55.34}{7.22}
& \pmstd{60.93}{3.30} & \pmstd{53.33}{2.52}
& \pmstd{62.43}{4.72} & \pmstd{43.69}{6.25} & \pmstd{47.24}{4.90} \\

\textit{w/o Early Stop}
& \pmstd{81.34}{3.74} & \pmstd{48.66}{7.91} & \pmstd{55.99}{6.18}
& \pmstd{66.31}{3.23} & \pmstd{82.77}{2.14}
& \pmstd{75.53}{4.09} & \pmstd{59.13}{6.12} & \pmstd{57.38}{5.22} \\

\textit{Leiden $\rightarrow$ Louvain}
& \pmstd{83.15}{4.84} & \pmstd{57.75}{9.06} & \pmstd{56.37}{8.02} 
& \pmstd{68.16}{3.30} & \pmstd{83.04}{2.06} 
& \pmstd{75.68}{4.02} & \pmstd{59.83}{6.18} & \pmstd{55.97}{4.81} \\

\bottomrule
\end{tabular}
\vspace{-0.2cm}
\captionsetup{font=footnotesize}
\caption{Ablation study of \mname using \textit{Qwen3-14B} as backbone LLM on MIMIC-III and MIMIC-IV datasets. All evaluation metrics are defined such that higher values indicate better performance ($\uparrow$). The best results are highlighted in \textbf{Bold}.}
\label{tab:ablation_mimic}
\vspace{-0.4cm}
\end{table*}
\noindent\textbf{Component Ablation.} 
We ablate each component of \mname{} on two EHR benchmarks 
using \textit{Qwen3-14B}; results are reported in 
Table~\ref{tab:ablation_mimic}. The three ablations directly 
verify the three challenges identified in 
Section~\ref{sec:pilot_study}.
\textit{(i) w/o EHR-emb} (\greennum{$-18.50$} F1): 
\textbf{generic semantic similarity is insufficient.} 
Replacing the EHR encoder with \textit{Qwen3-Embedding-8B} 
fails to capture clinically meaningful relationships 
required for effective graph-based search 
(\emph{Observation I}).
\textit{(ii) w/o Cohort Discovery} (\greennum{$-6.09$} AUPRC): 
\textbf{cohort-level retrieval matters.} Directly initializing 
anchors as the top-$K_a$ nearest patients reduces \mname{} 
to instance-level retrieval, which is vulnerable to noise 
from idiosyncratic cases (\emph{Observation II}). Replacing 
Leiden with Louvain~\cite{blondel2008fast} causes consistent 
drops, indicating that higher-quality partitions further 
benefit cohort retrieval and frontier expansion.
\textit{(iii) w/o Greedy \& Frontier Expansion} 
(\greennum{$-14.83$} AUPRC): \textbf{independent scoring fails.} 
Selecting demonstrations solely by individual $\Delta H$ 
scores implicitly assumes linear utility accumulation, 
ignoring redundancy and interaction effects that cause 
diminishing marginal gains (\emph{Observation III}). 
\textit{w/o Early Stop} further degrades performance by 
admitting demonstrations with negative marginal gains. \mname{} further generalizes across multiple pretrained 
EHR encoders, confirming its effectiveness is not tied to 
any specific representation 
(Appendix~\ref{app:encoder_generality}, 
Table~\ref{tab:encoder_gain}).

\begin{figure}[t]
    \centering
    \includegraphics[scale=0.14]{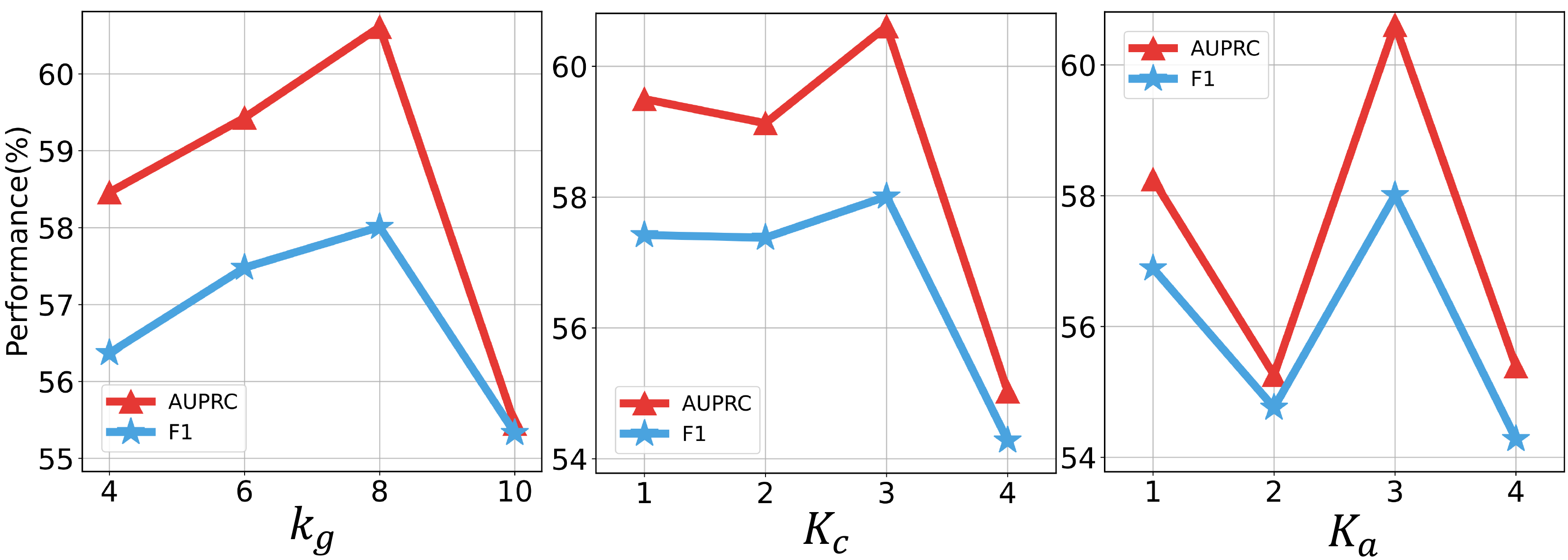}
    \vspace{-0.6cm}
    \captionsetup{font=footnotesize}
    \caption{Performance of \mname over different 
    hyper-parameters on MIMIC-IV.}
    \label{fig:hyper}
\vspace{-1.0em}
\end{figure}

\noindent\textbf{Hyper-parameter Sensitivity and Shot Scaling Analysis.} 
For each of the three key hyper-parameters (cohort count 
$K_c$, anchor count $K_a$, and graph neighborhood size 
$k_g$), \mname{} performs best at intermediate values and 
degrades when the parameter is set too small or too large 
(Appendix Figure~\ref{fig:hyper}), reflecting an 
interpretable coverage-noise trade-off rather than brittle 
sensitivity. Under varying demonstration budgets $K$, 
\textbf{\mname{} is the only method whose performance 
consistently improves with more demonstrations} 
(Figures~\ref{fig:scale_bar_part_mimic3} 
and~\ref{fig:scale_bar_part_mimic4_readmission}), uniquely 
sustaining a stable $\Delta H$ increase across shots 
(Figure~\ref{fig:entropy}) by composing demonstrations for 
joint information gain rather than scoring them 
independently. Baselines, in contrast, exhibit non-monotonic 
trends, confirming that the gain comes from \emph{how} 
demonstrations are selected and composed, not from 
\emph{how many}. Detailed analyses are in 
Appendix~\ref{app:Detailed_Hyper-Parameter_Analysis} 
and~\ref{app:shot_scaling}.

\noindent\textbf{Cost Analysis.} 
\mname{} incurs only \textbf{seconds-level additional inference 
overhead} per patient, while 
delivering substantial performance gains 
(Table~\ref{tab:main_results_mimic}). This favorable 
performance--latency trade-off is enabled by lazy greedy 
optimization with frontier expansion; the detailed runtime 
breakdown is provided in Appendix~\ref{app:runtime_analysis}.

\begin{figure}[t]
    \centering
    \begin{subfigure}[t]{0.47\columnwidth}
        \centering
        \includegraphics[width=\textwidth]{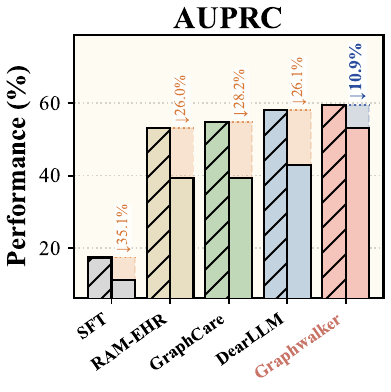}
    \end{subfigure}
    \hfill
    \begin{subfigure}[t]{0.47\columnwidth}
        \centering
        \includegraphics[width=\textwidth]{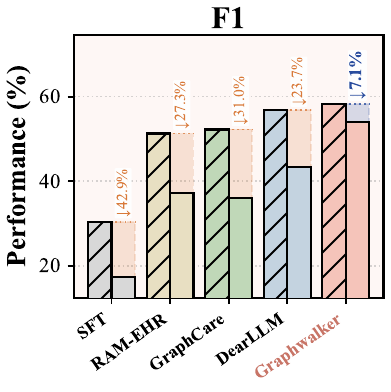}
    \end{subfigure}
    \vspace{-0.3cm}
    \captionsetup{font=footnotesize}
    \caption{In-domain (hatched) vs.\ cross-dataset OOD (solid) 
    on MIMIC-III mortality. Left: AUPRC; right: F1. Shaded 
    regions denote the relative drop. AUROC: 
    Appendix~\ref{app:supervised_comparison}.}
    \label{fig:supervised_comparison}
    \vspace{-0.4cm}
\end{figure}

\vspace{-0.6em}
\subsection{Compare to Supervised Methods (RQ3)}
\label{sec:rq3}
\vspace{-0.3em}
We evaluate \mname{} against 
representative supervised models on MIMIC-III mortality in 
two regimes: \textbf{(i) in-domain} (trained and tested on 
MIMIC-III), and \textbf{(ii) cross-dataset OOD} (trained on 
MIMIC-IV, tested on MIMIC-III without fine-tuning; \mname{}'s graph 
and retrieval base are built exclusively on MIMIC-IV's 
training split).  See 
Figure~\ref{fig:supervised_comparison} and 
Appendix~\ref{app:supervised_comparison} for details.

\noindent\textbf{Takeaway \ding{204}}: In-domain, \mname{} 
outperforms all four supervised baselines, including 
LLM-augmented hybrid pipelines such as 
RAM-EHR~\cite{xu2024ram}, GraphCare~\cite{jiang2024graphcare}, 
and DearLLM~\cite{xu2025dearllm}. Under cross-dataset distribution shift, supervised 
methods degrade notably (\greennum{15.4\%--42.9\%} drop), 
whereas \mname{} stays stable (\greennum{3.4\%--10.9\%} drop) 
and surpasses all baselines, reflecting a paradigm-level 
shift from source-specific feature correlations to reasoning 
by analogy over retrieved patient cases.

\vspace{-0.8em}
\subsection{Agent Skill Integration (RQ4)}
\label{sec:rq4}
\vspace{-0.3em}
Beyond standalone demonstration selection, we examine 
whether \mname{} can serve as a \textbf{composable 
patient-analogy skill}: a pluggable upstream module that 
supplies high-quality patient case context to downstream 
agentic workflows. To this end, we (i) implement a 
\textbf{black-box variant} of \mname{} that replaces 
entropy-based scoring with LLM self-evaluation, applicable 
to closed-source LLMs without access to logits; and (ii) 
integrate it with \textbf{ReAct}~\cite{yao2022react}, 
where the selected demonstrations remain unchanged but 
direct answer generation is replaced with ReAct-style 
reasoning. Experiments use MIMIC-III mortality; details in 
Appendix~\ref{app:blackbox_agentic_graphwalker}.

\noindent\textbf{Takeaway \ding{205}}: On closed-source LLMs, 
the black-box variant of \mname{} substantially outperforms 
random and similarity-based baselines, confirming that its 
advantage stems from cohort-guided, target-conditioned 
demonstration selection rather than any specific scoring 
interface. When combined with ReAct, \mname{} further 
improves across all three metrics whereas ReAct brings 
limited gains to weaker baselines, positioning \mname{} as 
a composable patient-analogy skill that complements 
agent-style reasoning.

\noindent\textbf{Extended Investigations and Key Insights.}
We further evaluate \mname{} on additional medical reasoning 
benchmarks spanning diverse task formats, where it achieves 
an average gain of \rednum{3.5\%$\uparrow$} 
(Table~\ref{tab:medqa_cmb_results}; 
Appendix~\ref{app:apply2more_task}), demonstrating its 
\textbf{generalizability to broader real-world medical 
reasoning scenarios}. A \textbf{case study} is also 
provided in Appendix~\ref{app:case_study}.

\vspace{-0.6em}
\section{Conclusion}
\vspace{-0.3em}
We presented \mname{}, an inference-time framework for 
clinical reasoning by patient analogy that addresses 
perspective limitation, cohort awareness, and information 
aggregation through dual-view selection, cohort discovery, 
and lazy greedy search with frontier expansion. Experiments 
across EHR and medical reasoning benchmarks demonstrate its 
effectiveness, positioning \mname{} as a pluggable 
patient-analogy skill for LLM-based clinical reasoning.

\section{Limitations}

Although \mname{} shows consistent gains across EHR prediction, additional medical reasoning benchmarks, and black-box or agentic extensions, several limitations remain.
Our current evaluation still focuses on structured EHR-based prediction and selected medical reasoning settings, partly due to computational constraints. Extending \mname{} to broader clinical scenarios, such as multimodal records and interactive clinical workflows, is an especially promising direction for future work, and we expect its core cohort-guided and target-conditioned demonstration selection mechanism to remain effective in such richer clinical environments.
Moreover, \mname{} is not restricted to the white-box few-shot setting: it can already be adapted to black-box LLMs and integrated with agentic reasoning scaffolds. While we have not yet systematically evaluated it in more complex clinical agent systems involving iterative tool use and evolving reasoning skills, this remains a particularly promising direction for future work, and our current black-box and ReAct-based results already provide encouraging preliminary evidence.

\section{Ethical Considerations}

Clinical decision-making is a complex and high-stakes process that relies on comprehensive medical evidence and professional judgment.
Although LLMs can assist clinicians by organizing information, highlighting relevant patterns, or supporting exploratory reasoning, there is a risk that their outputs may be misunderstood or inappropriately treated as definitive medical advice.
We emphasize that the system studied in this work is not intended to function as a diagnostic tool, nor to replace qualified healthcare professionals.
Any model-generated reasoning or suggestions should be interpreted only as auxiliary information and must be considered alongside established clinical records, examinations, and expert assessment.
All experiments in this study are conducted on publicly available and de-identified datasets, and no personally identifiable information or human subject data is involved.
We acknowledge that the underlying datasets may exhibit demographic imbalances, including disparities across race, ethnicity, age, and other population attributes, which could potentially affect model behavior and generalizability.
Our method is developed and evaluated solely for research purposes and is not deployed in real-world clinical settings.
While the proposed approach improves the reliability and coherence of in-context reasoning under limited information, any future clinical application would require extensive validation, careful safety evaluation, and strict compliance with medical regulations and ethical guidelines.
We view this work as a step toward more responsible and transparent medical AI systems, rather than a substitute for human clinical expertise.

\bibliography{custom}

\appendix

\newpage
\appendix

\section{Related Work}
\label{app:related_work}

Recent years have witnessed growing interest in applying LLMs 
to clinical reasoning over EHRs. One line of work adapts LLMs 
to the medical domain through continual pre-training or 
post-training on clinical 
corpora~\cite{zhang2025adept,ding20253ds,fang2025toward,ding2025promed}; 
a parallel line builds hybrid pipelines that distill 
LLM-derived knowledge into supervised EHR 
predictors~\cite{jiang2024graphcare,xu2025dearllm,xu2024ram}. Both 
directions, however, rely on parameter updates over 
task-specific labeled data. The former leaves LLMs limited in 
modeling longitudinal EHR data and activating clinically 
meaningful temporal patterns stored in parametric 
memory~\cite{brown2024not,chen2024clinicalbench,zhu2024clinicrealm}; 
the latter inherits the brittleness of its downstream 
supervised predictor to variations in feature order, 
availability, and encoding across heterogeneous EHR 
systems~\cite{fang2025toward}. These limitations motivate the 
exploration of inference-time adaptation strategies.

\paragraph{Demonstration Selection for ICL.}
In-context learning (ICL)~\cite{brown2020language} enables LLMs to adapt at inference time by conditioning on a small set of demonstrations, and has been widely studied in both general NLP and domain-specific settings.
Existing demonstration selection methods can be broadly grouped into two paradigms.
Data-perspective approaches retrieve demonstrations based on embedding similarity, assuming that representation-level proximity correlates with reasoning utility~\cite{liu2022makes,hongjin2022selective,robertson2009probabilistic,yu2024smart}.
Model-perspective approaches instead leverage LLM-internal signals, such as perplexity, conditional entropy, gradients, or influence estimates, to assess the usefulness of individual demonstrations~\cite{gonen2023demystifying,peng2024revisiting,liumakes,li2025delta,zhang2025linear,qin2024context}.
However, existing demonstration selection methods largely lack a principled framework that jointly incorporates clinical knowledge and model-aware reasoning signals for EHR-based clinical reasoning.

\section{Experimental Details}\label{app:exp-details}

\subsection{Dataset Descriptions}
\label{app:dataset}
In this section, we provide a detailed introduction of the datasets used in our study, along with the preprocessing procedures applied to the structured EHR data.

We conduct experiments on two real-world EHR benchmarks: 
\textbf{MIMIC-III v1.4}~\cite{johnson2016mimic} and 
\textbf{MIMIC-IV v2.2}~\cite{johnson2023mimic}. Both are 
large-scale, publicly available critical-care databases 
collected at the Beth Israel Deaconess Medical Center, providing 
longitudinal, de-identified ICU records with heterogeneous 
clinical events (e.g., vital signs, laboratory tests, and 
medication administrations). The resulting irregular, 
long-horizon patient trajectories make these datasets 
well-suited for evaluating clinical reasoning over EHRs.

\paragraph{Data Preprocessing.} 
We preprocess both datasets following the standardized 
PyEHR~\cite{zhu2024pyehr} pipeline, which is also adopted by 
recent EHR benchmarks~\cite{gao2024comprehensive}. Time-stamped 
clinical events within each ICU stay are temporally aggregated 
into daily records; for long stays, we retain the most recent 
seven days, which prior work has shown to capture the most 
clinically salient trajectories. Missing values are handled 
using the Last Observation Carried Forward (LOCF) 
strategy~\cite{wells2013strategies}.

For each patient, we use \emph{all historical visits except the 
last one} as input and predict the outcome associated with the 
last visit. This setup strictly excludes the outcome visit from 
the input, ensuring that no information from the prediction 
target is leaked into the input features. Specifically, for 
\textbf{in-hospital mortality}, we predict whether the patient 
dies during the last visit; for \textbf{length-of-stay (LOS)}, 
we predict the LOS bin~\cite{harutyunyan2019multitask} of the 
last visit; for \textbf{ICU readmission}, we predict whether a 
new ICU admission occurs within 30 days of the last visit's 
discharge. The same prediction protocol is applied uniformly to 
all baselines and to \mname{}.

To enable controlled comparison across a broad set of LLM 
backbones and case selection baselines, each requiring 
inference-time forward passes per patient, we follow prior EHR 
studies~\cite{zhu2026clinicrealm} and split each dataset into 
training, validation, and test sets at an 8:1:1 ratio, with 
the test set fixed across all experiments to ensure fair 
comparison.

\subsection{Tasks}
\label{app:task}
\noindent\textbf{Definition 1 (EHR Dataset).}  
A patient's EHR is represented as a sequence of $T$ time-ordered visits $\mathbf{X} = [\mathbf{x}_1, \mathbf{x}_2, \cdots, \mathbf{x}_T]$, where each visit $\mathbf{x}_t = \{l_{t,1}, l_{t,2}, \cdots, l_{t,n_t}\}$ contains $n_t$ lab test features.

\noindent\textbf{Definition 2 (Mortality Prediction).}  
Given $\mathbf{X}$, predict whether the patient will survive the hospital stay. The label $y \in \{0,1\}$ denotes death ($y = 1$) or survival ($y = 0$).

\noindent\textbf{Definition 3 (Readmission Prediction).}  
Given $\mathbf{X}$, predict whether the patient will be readmitted within 30 days after discharge. The label $y \in \{0,1\}$ indicates readmission ($y = 1$) or not ($y = 0$).

\noindent\textbf{Definition 4 (Length-of-Stay Prediction).}  
Given $\mathbf{X}$, predict the length of the patient’s hospital stay. 
Following prior work~\cite{harutyunyan2019multitask}, we formulate this task as a classification problem, where the label $y \in \{1, 2, \cdots, C\}$ indicates the length-of-stay category corresponding to predefined time intervals.

\subsection{Baseline Implementation Details}
\label{appen:baseline_details}
We compare \mname with representative baselines spanning multiple paradigms. 
In this subsection, we provide implementation details for each baseline.
For all methods, we strictly constrain the LLM to directly output the final prediction in the task-specific format, without generating any intermediate reasoning or explanations, following the prompt templates described in Appendix~\ref{app:prompt}.

\begin{itemize}[leftmargin=*,label=\textbullet,noitemsep,topsep=2pt]
    \item \textbf{Vanilla ICL baselines.}  
    \textit{Zero-shot} corresponds to a special case of ICL where no demonstration examples are provided, and the LLM directly produces predictions based on the task description and patient EHR data. 
    \textit{Random} samples a fixed number of patient examples uniformly at random as demonstrations.
    Both baselines adopt the same prompt structure as described in Appendix~\ref{app:prompt}, differing only in whether in-context examples are included.

    \item \textbf{Supervised.}  
    \textit{SFT} adapts LLMs to downstream tasks by fine-tuning model parameters on task-specific data.
    Specifically, we organize the training set into question--answer pairs using the same prompt templates described in Appendix~\ref{app:prompt}, and apply parameter-efficient fine-tuning via LoRA~\cite{hu2022lora} to update the backbone LLM.

    \item \textbf{Data Perspective Selection.}  
    \textit{Semantic-emb} retrieves top-$k$ demonstrations based on similarity in the embedding space of \textit{Qwen3-Embedding-8B}~\cite{zhang2025qwen3}, a widely used and strong general-purpose semantic model. For efficiency, we compute embeddings using vLLM~\cite{kwon2023efficient} to accelerate large-scale embedding inference.
    \textit{EHR-emb} retrieves top-$k$ demonstrations based on similarity computed from embeddings generated by \textit{SMART}~\cite{yu2024smart}, a pretrained model specifically designed for EHR data.
    We follow the hyperparameter settings recommended in the original work and use the \texttt{[CLS]} token embedding as the patient representation, as it encodes global semantic information of the longitudinal EHR sequence.
    \textit{Time-series}~\cite{kawarada2024demonstration} retrieves top-$k$ demonstrations based on the similarity between the test patient and training examples’ longitudinal EHR time series.  
    We use a scale-invariant masked cosine similarity by aligning overlapping prefixes, truncating to the minimum length, and computing cosine similarity only over positions with valid values, ignoring missing entries.  
    Demonstrations are ranked by similarity, and the top-$k$ are selected as in-context examples.

    \item \textbf{Model Perspective Selection.}  
    \textit{SPELL}~\cite{gonen2023demystifying} selects demonstrations by ranking candidates with LM perplexity (i.e., teacher-forced next-token cross-entropy) and choosing the lowest-perplexity top-$k$ examples. In our implementation, we score all training examples with the same causal LM and select the top-$k$ demonstrations by sorting the average loss in ascending order. The resulting few-shot set is shared across all test instances and inserted into the ICL prompt as demonstrations, without per-test-instance re-selection.
    
    \textit{Influence}~\cite{nguyen2023context} estimates the importance of demonstrations by measuring their marginal contribution to the model's performance on a validation set. It calculates influence scores based on the average performance difference between sampled subsets that include a specific example and those that exclude it.
    
    \textit{IDS}~\cite{qin2024context} selects demonstrations by calculating the semantic similarity between training examples and reasoning paths generated through Zero-shot CoT. The framework iteratively refines the demonstration set by utilizing the model's self-generated reasoning from the previous round as an updated query for subsequent retrieval.
    
    \textit{CONE}~\cite{peng2024revisiting} selects demonstrations by minimizing the conditional entropy of the test input on a per-example basis, without modeling complementarity among selected examples.
    Following the original implementation, we first perform a semantic pre-filtering step using \textit{Qwen3-Embedding-8B} to retrieve the top-10 semantically similar candidates.
    We then select the top-$k$ demonstrations with the largest reduction in conditional entropy, where the entropy is computed independently for each candidate example.

    \textit{LMS3}~\cite{liumakes} selects demonstrations by balancing LLM-oriented semantic similarity and inference stability, derived from a theoretical bound on reasoning efficacy. It uniquely incorporates a demonstration rejection mechanism that adaptively filters out unsuitable examples to prevent negative transfer.
    
    \textit{Delta-KNN}~\cite{li2025delta} selects demonstrations by prioritizing candidates that were empirically observed to be most helpful for similar inputs, using a cached example-to-example gain matrix. In our implementation, we first take a fixed training subset of examples to construct the Delta matrix via paired zero-shot vs.\ one-shot inference, and cache the resulting matrix for each dataset/LLM setting. At test time, we use \textit{Qwen3-Embedding-8B} to retrieve the top-10 semantically similar candidates, then re-rank them with the Delta-based scores and select the final top-$k$ demonstrations. The selected three demonstrations are re-selected per test instance and inserted into the ICL prompt as demonstrations for final inference.

    \textit{GradSel}~\cite{zhang2025linear} estimates the utility of sampled demonstration subsets using a first-order approximation in the input-embedding space, based on the model output and gradients computed at anchor prompts. It then aggregates subset-level scores into point-wise demonstration scores and selects the top-ranked examples as final demonstrations.

\end{itemize}

\subsection{Implementation Details of \mname}
\label{appendix:Implementation}

For patient graph construction, we set the number of nearest neighbors to $k_g = 8$.
For instance-aware cohort retrieval, the number of retrieved cohorts is set to $K_c = 3$, and within each retrieved cohort, we select $K_a = 3$ anchor nodes to initialize the search frontier.
A detailed analysis and discussion of these key hyperparameters are provided in Appendix~\ref{app:Detailed_Hyper-Parameter_Analysis}.
For graph-based cohort discovery, the resolution parameter of the Leiden algorithm is set to $0.9$. All experiments are repeated three times with different random seeds, and we report the mean performance along with the standard deviation.
To accelerate the computation of conditional entropy during LLM-guided demonstration selection, we adopt \texttt{vLLM}~\cite{kwon2023efficient} for efficient inference.

\section{Additional Experiment Results}\label{app:results}

\begin{table*}[!ht]
\centering
\fontsize{7.5pt}{8pt}\selectfont
\setlength{\tabcolsep}{4.0pt}
\renewcommand{\arraystretch}{1.3}

\begin{tabular}{l|l|ccc|cc|ccc}
\toprule
\multicolumn{2}{c|}{\textbf{Method}}
& \multicolumn{3}{c|}{\textbf{MIMIC-III Mortality}}
& \multicolumn{2}{c|}{\textbf{MIMIC-III LOS}}
& \multicolumn{3}{c}{\textbf{MIMIC-IV Readmission}} \\
\textbf{Paradigm} & \textbf{Approach}
& AUROC $\uparrow$ & AUPRC $\uparrow$ & F1 $\uparrow$
& ma-ROC $\uparrow$ & mi-ROC $\uparrow$
& AUROC $\uparrow$ & AUPRC $\uparrow$ & F1 $\uparrow$ \\
\midrule

\rowcolor{gray!10}
\multicolumn{10}{c}{\textit{Qwen3-32B}} \\
\midrule
\multirow{2}{*}{Vanilla} & Zero-shot
& \pmstd{65.67}{5.75} & \pmstd{30.54}{7.56} & \pmstd{38.43}{6.99}
& \pmstd{55.79}{3.33} & \pmstd{23.53}{2.46}
& \pmstd{44.19}{4.57} & \pmstd{25.06}{3.88} & \pmstd{42.57}{3.83} \\
& Random
& \pmstd{59.90}{5.06} & \pmstd{24.63}{5.29} & \pmstd{36.22}{4.94}
& \pmstd{59.50}{3.51} & \underline{\pmstd{80.44}{2.00}}
& \pmstd{48.25}{4.65} & \pmstd{28.13}{4.51} & \pmstd{42.22}{3.77} \\
\midrule
\multirow{3}{*}{Data Perspective} & Semantic-emb
& \pmstd{60.62}{5.80} & \pmstd{19.87}{4.61} & \pmstd{32.08}{5.59}
& \pmstd{57.65}{3.65} & \pmstd{74.16}{2.38}
& \pmstd{47.86}{4.71} & \pmstd{26.07}{3.89} & \pmstd{42.47}{3.85} \\
& EHR-emb
& \underline{\pmstd{85.02}{3.21}} & \underline{\pmstd{58.56}{8.02}} & \underline{\pmstd{58.84}{6.01}}
& \underline{\pmstd{63.02}{3.33}} & \pmstd{67.00}{2.71}
& \underline{\pmstd{61.32}{4.45}} & \underline{\pmstd{44.25}{5.92}} & \underline{\pmstd{46.24}{4.06}} \\
& Time-series
& \pmstd{63.71}{4.93} & \pmstd{28.74}{6.28} & \pmstd{37.58}{5.29}
& \pmstd{51.95}{4.04} & \pmstd{38.06}{2.98}
& \pmstd{52.34}{4.78} & \pmstd{27.02}{4.39} & \pmstd{40.79}{3.97} \\
\midrule
\multirow{7}{*}{Model Perspective} & SPELL
& \pmstd{63.26}{4.98} & \pmstd{27.26}{5.64} & \pmstd{38.33}{5.45}
& \pmstd{53.87}{3.76} & \pmstd{75.06}{2.27}
& \pmstd{48.56}{4.77} & \pmstd{25.90}{4.11} & \pmstd{40.17}{3.86} \\
& Influence
& \pmstd{62.45}{5.55} & \pmstd{29.28}{6.41} & \pmstd{40.18}{6.30}
& \pmstd{54.24}{3.02} & \pmstd{41.07}{1.56}
& \pmstd{50.45}{3.49} & \pmstd{27.63}{3.42} & \pmstd{42.69}{3.94} \\
& IDS
& \pmstd{56.72}{5.64} & \pmstd{28.37}{6.91} & \pmstd{35.23}{5.03}
& \pmstd{52.47}{3.12} & \pmstd{45.82}{2.47}
& \pmstd{46.93}{4.66} & \pmstd{27.71}{4.34} & \pmstd{43.16}{3.79} \\
& CONE
& \pmstd{64.28}{6.59} & \pmstd{27.83}{7.35} & \pmstd{36.05}{6.90}
& \pmstd{56.30}{3.73} & \pmstd{68.90}{2.64}
& \pmstd{45.35}{4.83} & \pmstd{27.85}{4.77} & \pmstd{41.73}{3.79} \\
& LMS3
& \pmstd{62.03}{5.12} & \pmstd{25.14}{5.03} & \pmstd{36.68}{4.79}
& \pmstd{58.14}{3.73} & \pmstd{73.15}{2.55}
& \pmstd{45.28}{4.40} & \pmstd{26.09}{4.01} & \pmstd{42.79}{3.82} \\
& Delta-KNN
& \pmstd{61.08}{5.83} & \pmstd{29.41}{6.34} & \pmstd{39.36}{5.92}
& \pmstd{56.55}{4.09} & \pmstd{64.86}{2.64}
& \pmstd{51.75}{4.68} & \pmstd{26.74}{4.26} & \pmstd{40.37}{3.90} \\
& GradSel
& \pmstd{60.77}{5.22} & \pmstd{24.53}{4.77} & \pmstd{39.30}{5.68}
& \pmstd{53.87}{3.75} & \pmstd{44.22}{2.55}
& \pmstd{47.08}{4.02} & \pmstd{25.53}{3.41} & \pmstd{41.93}{3.80} \\
\midrule
\rowcolor[HTML]{F0F6FF}
Ours & \textit{GraphWalker}
& \textbf{\pmstd{85.88}{3.16}} & \textbf{\pmstd{59.19}{8.04}} & \textbf{\pmstd{61.89}{6.07}}
& \textbf{\pmstd{67.40}{3.51}} & \textbf{\pmstd{83.60}{2.34}}
& \textbf{\pmstd{77.57}{4.01}} & \textbf{\pmstd{63.09}{6.27}} & \textbf{\pmstd{59.51}{4.46}} \\
\bottomrule
\end{tabular}
\captionsetup{font=footnotesize}
\caption{Additional results on MIMIC-III and MIMIC-IV 
datasets using \textit{Qwen3-32B} as the backbone LLM. All 
metrics are higher-is-better ($\uparrow$); values are 
mean$\pm$std over three random seeds. \textbf{Bold}: best; 
\underline{underlined}: second-best.}
\label{tab:appendix_qwen32b_results}
\vspace{-0.4cm}
\end{table*}
\subsection{RQ1 Results}
\label{app:qwen32b_results}
To further evaluate the robustness of \mname under larger backbone models, we additionally report results using \textit{Qwen3-32B}. As shown in Table~\ref{tab:appendix_qwen32b_results}, \mname consistently achieves the best overall performance across all three clinical prediction tasks, including MIMIC-III Mortality, MIMIC-III LOS, and MIMIC-IV Readmission. 


\subsection{RQ2 Results}


\subsubsection{Effect of Different EHR Encoders}
\label{app:encoder_generality}
\begin{table*}[t!]
\centering
\fontsize{8pt}{9pt}\selectfont
\setlength{\tabcolsep}{6pt}
\renewcommand{\arraystretch}{1.15}
\begin{tabular}{l|ccc|ccc}
\toprule
\rowcolor{gray!10}
\textbf{Method}
& \multicolumn{3}{c|}{\textbf{MIMIC-III Mortality}}
& \multicolumn{3}{c}{\textbf{MIMIC-IV Readmission}} \\
\rowcolor{gray!10}
& AUROC$\uparrow$ & AUPRC$\uparrow$ & F1$\uparrow$
& AUROC$\uparrow$ & AUPRC$\uparrow$ & F1$\uparrow$ \\
\midrule
SMART (fine-tuned predictor)
& 80.64 & 45.40 & 47.18
& 74.41 & 54.62 & 57.38 \\
\mname{} (with SMART encoder)
& \textbf{84.19} & \textbf{59.46} & \textbf{58.24}
& \textbf{75.91} & \textbf{60.61} & \textbf{58.01} \\
\textit{$\Delta$}
& +3.55 & +14.06 & +11.06
& +1.50 & +5.99 & +0.63 \\
\midrule
ConCare (fine-tuned predictor)
& 69.19 & 36.01 & 40.65
& 70.30 & 41.87 & 50.26 \\
\mname{} (with ConCare encoder)
& \textbf{70.46} & \textbf{36.93} & \textbf{45.90}
& \textbf{77.01} & \textbf{59.49} & \textbf{59.44} \\
\textit{$\Delta$}
& +1.27 & +0.92 & +5.25
& +6.71 & +17.62 & +9.18 \\
\midrule
AdaCare (fine-tuned predictor)
& 60.67 & 19.37 & 26.81
& 78.40 & 59.37 & 54.87 \\
\mname{} (with AdaCare encoder)
& \textbf{65.61} & \textbf{27.13} & \textbf{37.89}
& \textbf{80.45} & \textbf{62.72} & \textbf{60.04} \\
\textit{$\Delta$}
& +4.94 & +7.76 & +11.08
& +2.05 & +3.35 & +5.17 \\
\bottomrule
\end{tabular}
\captionsetup{font=footnotesize}
\caption{Performance of \mname{} with different EHR encoders 
on MIMIC-III mortality and MIMIC-IV readmission, using 
\textit{Qwen3-14B} as the backbone LLM. For each encoder, we 
compare two settings: (i) the encoder fine-tuned end-to-end 
as a task-specific supervised predictor, and (ii) \mname{} 
using the same pretrained encoder (without fine-tuning) for 
graph-based demonstration selection. $\Delta$ denotes the 
absolute improvement (in percentage points) of \mname{} over 
the supervised fine-tuned baseline. Note that this differs 
from \textit{EHR-emb} in 
Table~\ref{tab:main_results_mimic}, which uses SMART 
embeddings for ICL retrieval without fine-tuning.}
\label{tab:encoder_gain}
\vspace{-0.4cm}
\end{table*}

\paragraph{Setup.} 
To assess the robustness of \mname{} to the choice of EHR 
representation, we instantiate it with three pretrained EHR 
encoders of different architectures: 
\textbf{SMART}~\cite{yu2024smart}, 
\textbf{ConCare}~\cite{ma2020concare}, and 
\textbf{AdaCare}~\cite{ma2020adacare}. For each encoder, we 
compare two settings: (i) the encoder fine-tuned end-to-end 
as a task-specific supervised predictor (\textit{encoder 
baseline}), and (ii) \mname{} using the same pretrained 
encoder \emph{without fine-tuning} for graph-based 
demonstration selection. Note that the encoder baseline here 
differs from \textit{EHR-emb} in 
Table~\ref{tab:main_results_mimic}, which uses SMART 
embeddings for demonstration retrieval without fine-tuning.

\paragraph{Results.} 
Table~\ref{tab:encoder_gain} shows that \mname{} consistently 
improves over all encoder baselines across both MIMIC-III 
mortality and MIMIC-IV readmission, with consistent absolute 
gains across metrics and encoders. The improvements hold 
across encoders of distinct architectural styles, indicating 
that \mname{}'s effectiveness arises from its demonstration 
selection mechanism rather than from any specific EHR 
representation. Notably, \mname{} surpasses these encoders 
\emph{even when they are fine-tuned end-to-end on the target 
task}, underscoring that the gains come from how 
demonstrations are selected and composed, not merely from 
the underlying encoder.

\subsubsection{Detailed Hyper-Parameter Analysis}
\label{app:Detailed_Hyper-Parameter_Analysis}

We analyze three key hyper-parameters of \mname{} on the 
validation set of \textit{MIMIC-IV Readmission} with 
\textit{Qwen3-14B} as the backbone LLM: the graph 
neighborhood size $k_g$, the number of retrieved cohorts 
$K_c$, and the number of anchor nodes per cohort $K_a$. 
All other settings are kept fixed.

Overall, \mname{} performs best under \emph{moderate} 
settings that balance candidate diversity and clinical 
coherence. Across all three hyper-parameters, performance 
remains strongest within a small intermediate range before 
declining when the candidate space becomes overly sparse or 
overly noisy, reflecting an interpretable coverage--noise 
trade-off rather than brittle sensitivity to narrow tuning.

\paragraph{Graph Neighborhood Size $k_g$.}
We vary $k_g$ from 4 to 10 (Figure~\ref{fig:hyper}(a)). 
Performance steadily improves as $k_g$ grows from 4 to 8, 
indicating that a moderate neighborhood captures richer 
population-level structure. When $k_g$ further increases to 
10, performance declines as the graph becomes overly dense 
and introduces weakly related neighbors, blurring local 
structure and propagating less informative signals during 
frontier expansion.

\paragraph{Number of Retrieved Cohorts $K_c$.}
Varying $K_c$ from 1 to 4 (Figure~\ref{fig:hyper}(b)), both 
AUPRC and F1 steadily improve from 1 to 3, suggesting that 
retrieving multiple cohorts enriches the candidate pool 
with diverse but clinically related subpopulations and 
mitigates over-specialization to a single cohort. 
Performance declines at $K_c = 4$, where additional 
loosely related cohorts begin to dilute clinically 
meaningful signals.

\paragraph{Anchor Nodes per Cohort $K_a$.}
Performance peaks at $K_a = 3$ (Figure~\ref{fig:hyper}(c)). 
A moderate number of anchors balances intra-cohort coverage 
and candidate quality—sufficient to represent diverse yet 
clinically consistent profiles within each cohort. Larger 
$K_a$ over-expands the initial frontier with marginal 
anchors, increasing redundancy and reducing the 
effectiveness of interaction-aware demonstration selection.

\subsubsection{Shot Scaling Analysis}
\label{app:shot_scaling}

\begin{figure*}[t]
    \centering
    \includegraphics[width=0.92\textwidth]{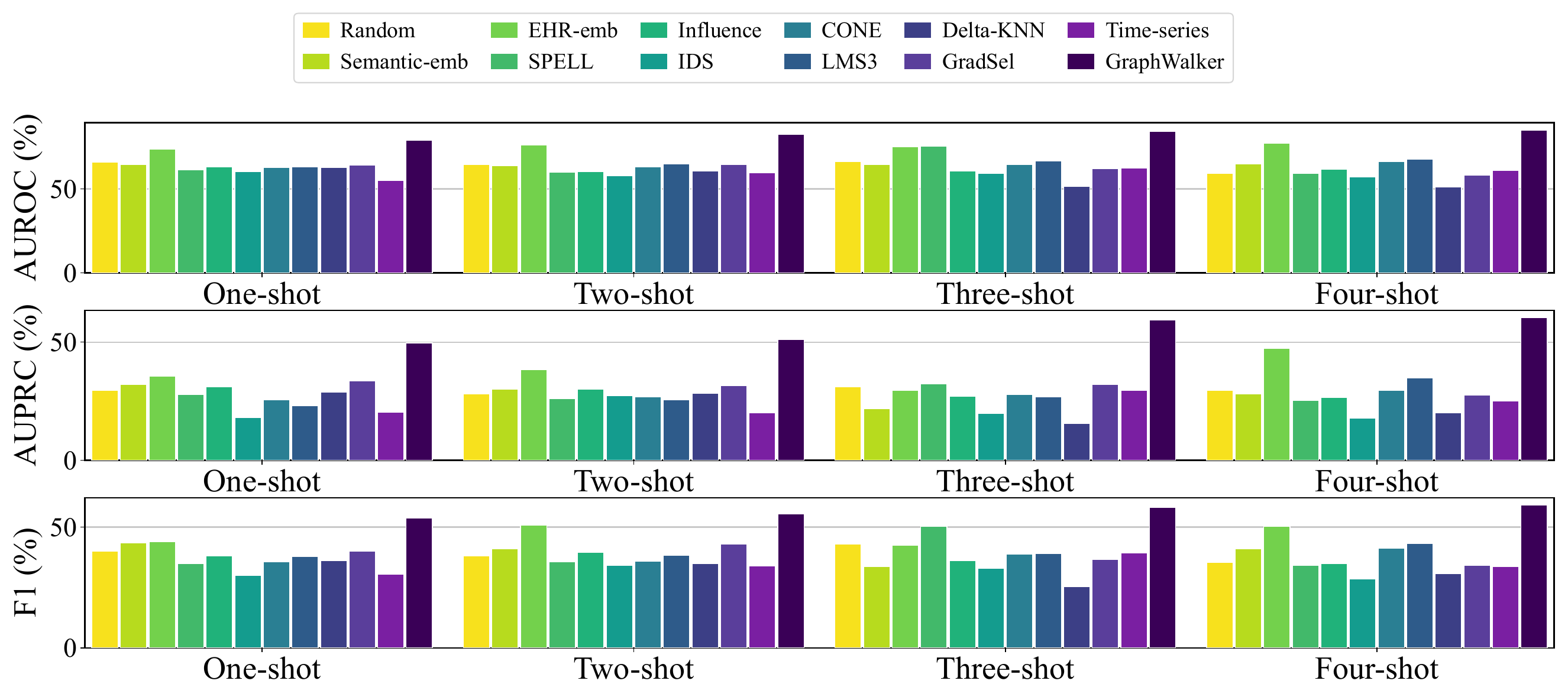}
\captionsetup{font=footnotesize}
    \caption{Shot scaling performance on MIMIC-III Mortality 
    with Qwen3-14B as the backbone LLM.}
    \label{fig:scale_bar_part_mimic3}
\end{figure*}

\begin{figure*}[t]
    \centering
    \includegraphics[width=0.92\textwidth]{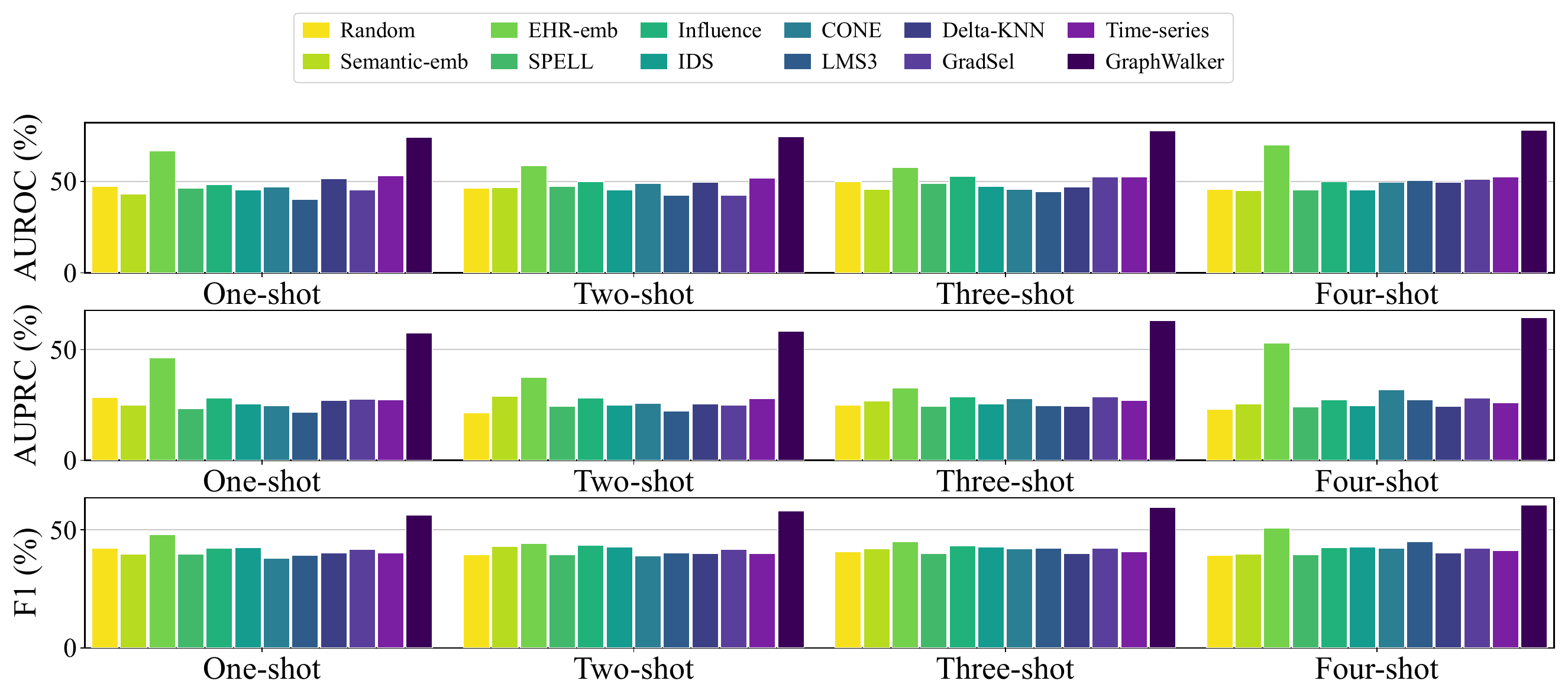}
    \captionsetup{font=footnotesize}
    \caption{Shot scaling performance on MIMIC-IV 
    Readmission with \textit{Qwen3-14B} as the backbone LLM.}
    \label{fig:scale_bar_part_mimic4_readmission}
\end{figure*}

To examine the effect of shot number on \mname{}, we 
evaluate its performance under varying $k$-shot settings 
($k \in \{1,2,3,4\}$) on \textit{MIMIC-III Mortality} and 
\textit{MIMIC-IV Readmission} with \textit{Qwen3-14B} as 
the backbone LLM (Figures~\ref{fig:scale_bar_part_mimic3} 
and~\ref{fig:scale_bar_part_mimic4_readmission}).

\paragraph{\mname{} consistently outperforms all baselines 
across $k$-shot settings.}
This demonstrates strong robustness to the choice of shot 
number. In contrast, baseline methods exhibit non-monotonic 
and dataset-dependent trends, indicating that 
\textbf{simply adding more demonstrations does not 
necessarily improve EHR reasoning performance}---irrelevant 
or misleading demonstrations can disrupt rather than enhance 
coherent reasoning.

\paragraph{\mname{} is the only method whose performance 
improves consistently with $k$.}
By grounding demonstration selection in cohort-level 
clinical similarity and applying LLM-guided composition-aware 
search, \mname{} progressively identifies demonstration sets 
that maximize information gain rather than adding redundant 
or harmful cases.

\paragraph{Purely model-perspective methods struggle 
relative to encoder-based baselines.}
Methods such as CONE, LMS3, and SPELL consistently 
underperform \textit{EHR-emb} and \mname{}, suggesting 
that \textbf{LLM-internal signals alone are insufficient 
to construct reliable demonstration sets for EHR reasoning} 
without grounding in clinically meaningful patient 
distributions.

\subsubsection{Runtime Analysis}
\label{app:runtime_analysis}

\begin{figure*}[t]
    \centering
    \includegraphics[width=\textwidth]{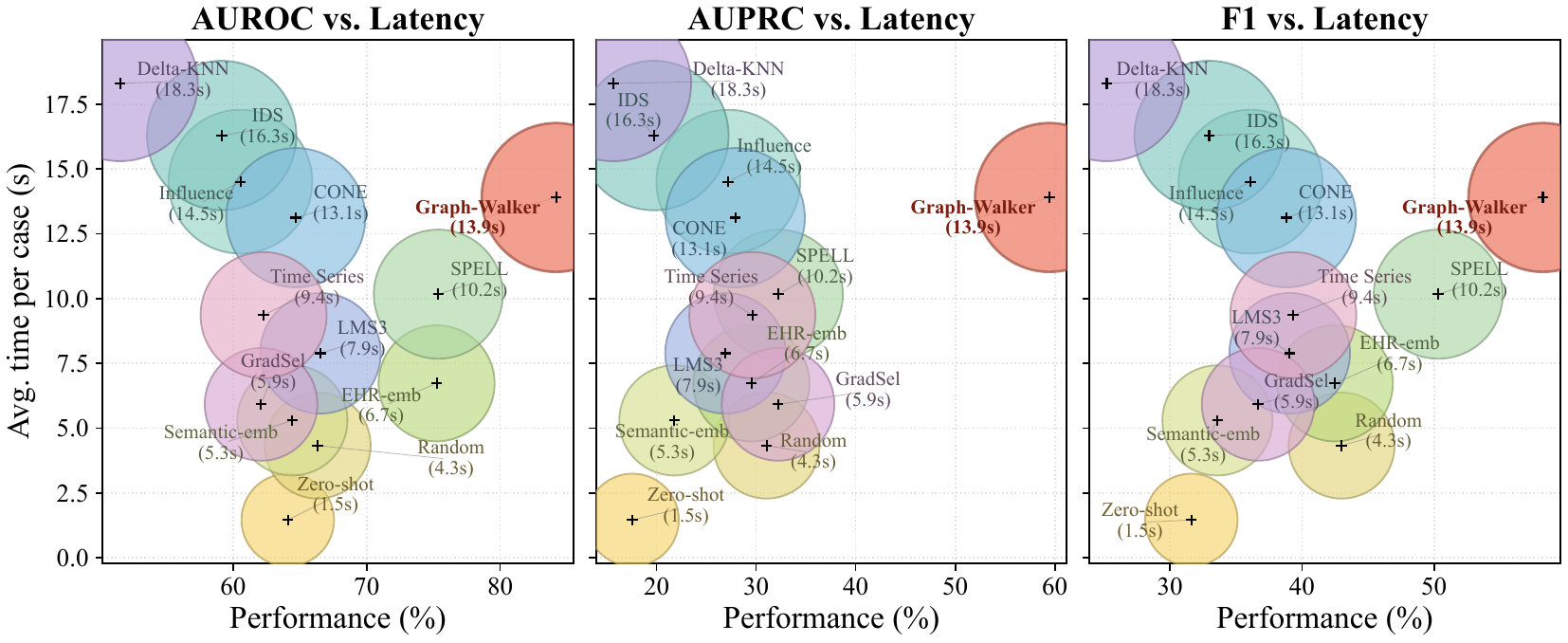}
    \vspace{-0.6cm}
    \captionsetup{font=footnotesize}
    \caption{Performance--latency tradeoff on MIMIC-III mortality 
    across AUROC, AUPRC, and F1 (left to right). Each bubble represents 
    one case retrieval method, with position showing mean performance 
    (x-axis) and average inference time per case (y-axis); bubble 
    color identifies the method. \mname{} (red) achieves the strongest 
    performance across all three metrics while keeping latency comparable 
    to prior LLM-based case selection methods.}
    \label{fig:latency_tradeoff}
    \vspace{-0.4cm}
\end{figure*}

The central design objective of \mname is to improve clinical reasoning accuracy by deliberately allocating a small additional inference budget at test time. In clinical decision-support scenarios, predictive accuracy often carries far greater practical value than marginal reductions in latency, since higher-quality predictions can directly affect the quality of downstream clinical decisions. Motivated by this, \mname invests extra inference cost to construct demonstration sets that are clinically coherent, complementary, and informative for each target patient---following the same high-level principle as test-time scaling in reasoning models, where additional computation is intentionally exchanged for better reasoning outcomes.

Crucially, this extra cost is tightly bounded rather than spent on unconstrained global search. Cohort retrieval, anchor initialization, frontier expansion, lazy greedy selection, and early stopping jointly restrict the inference budget to a small set of clinically relevant candidates, making \mname a targeted form of test-time demonstration optimization.

\paragraph{Experiments.}
We conduct runtime analysis on \textit{MIMIC-III Mortality} using \textit{Qwen3-14B} as the backbone LLM. For each demonstration selection method, we measure the average wall-clock evaluation time per test case---including both demonstration selection and final prediction generation---and compare the resulting performance--latency tradeoff across methods.

\paragraph{Results.}
As shown in Figure~\ref{fig:latency_tradeoff}, \mname achieves the strongest performance across all three metrics while keeping inference cost on par with existing LLM-based selection methods. The search overhead is effectively bounded by the cohort-guided frontier design rather than expanded into exhaustive global exploration. Two observations make this concrete:

\noindent\textit{(i) The absolute overhead is modest.} Compared with the strongest similarity-based baseline \textit{EHR-emb}, \mname incurs only $7.2$s additional inference time per case ($13.9$s vs.\ $6.7$s), yet yields substantial gains of $+9.0$ AUROC, $+29.9$ AUPRC, and $+15.8$ F1.

\noindent\textit{(ii) The overhead is competitive among LLM-based selection methods.} Among approaches that similarly invoke LLM-side scoring, \mname ($13.9$s) is on par with \textit{CONE} ($13.1$s), and notably faster than \textit{Influence} ($14.5$s), \textit{IDS} ($16.3$s), and \textit{Delta-KNN} ($18.3$s), while delivering substantially larger accuracy gains across all three metrics.

Together, these results show that \textbf{\mname converts a limited test-time cost into significant improvements in clinical prediction quality.}

\subsection{RQ3 Results}
\label{app:supervised_comparison}

\begin{table*}[t!]
\centering
\fontsize{6.5pt}{7.5pt}\selectfont
\setlength{\tabcolsep}{2.5pt}
\renewcommand{\arraystretch}{1.1}
\begin{tabular}{l|ccc|ccc|ccc}
\toprule
\rowcolor{gray!10}
\textbf{Method}
& \multicolumn{3}{c|}{\textbf{AUROC}}
& \multicolumn{3}{c|}{\textbf{AUPRC}}
& \multicolumn{3}{c}{\textbf{F1}} \\
\rowcolor{gray!10}
& In-Dom. & OOD & Drop 
& In-Dom. & OOD & Drop 
& In-Dom. & OOD & Drop \\
\midrule
SFT 
& \pmstd{68.34}{6.29} & \pmstd{51.04}{5.47} & \greennum{$\downarrow$25.3\%}
& \pmstd{17.48}{4.88} & \pmstd{11.34}{4.11} & \greennum{$\downarrow$35.1\%}
& \pmstd{30.43}{6.35} & \pmstd{17.38}{5.12} & \greennum{$\downarrow$42.9\%} \\
RAM-EHR~\cite{xu2024ram}
& \pmstd{80.58}{3.48} & \pmstd{59.26}{3.35} & \greennum{$\downarrow$26.5\%}
& \pmstd{53.02}{3.87} & \pmstd{39.23}{4.47} & \greennum{$\downarrow$26.0\%}
& \pmstd{51.26}{3.44} & \pmstd{37.26}{3.67} & \greennum{$\downarrow$27.3\%} \\
GraphCare~\cite{jiang2024graphcare}
& \pmstd{80.21}{3.55} & \pmstd{62.34}{5.57} & \greennum{$\downarrow$22.3\%}
& \pmstd{54.77}{3.19} & \pmstd{39.35}{3.67} & \greennum{$\downarrow$28.2\%}
& \pmstd{52.22}{3.75} & \pmstd{36.04}{4.36} & \greennum{$\downarrow$31.0\%} \\
DearLLM~\cite{xu2025dearllm}
& \pmstd{82.77}{3.63} & \pmstd{69.99}{5.21} & \greennum{$\downarrow$15.4\%}
& \pmstd{58.02}{3.24} & \pmstd{42.85}{6.39} & \greennum{$\downarrow$26.1\%}
& \pmstd{56.79}{3.75} & \pmstd{43.32}{6.06} & \greennum{$\downarrow$23.7\%} \\
\midrule
\rowcolor{blue!5}
\textbf{\mname{}}
& \textbf{\pmstd{84.19}{3.35}} & \textbf{\pmstd{81.29}{3.91}} & \textbf{\rednum{$\downarrow$3.4\%}}
& \textbf{\pmstd{59.46}{7.64}} & \textbf{\pmstd{52.98}{6.01}} & \textbf{\rednum{$\downarrow$10.9\%}}
& \textbf{\pmstd{58.24}{5.89}} & \textbf{\pmstd{54.08}{5.84}} & \textbf{\rednum{$\downarrow$7.1\%}} \\
\bottomrule
\end{tabular}
\captionsetup{font=footnotesize}
\caption{Performance comparison of \mname{} against supervised 
methods on MIMIC-III in-hospital mortality. ``In-Dom.'' refers 
to training and testing on MIMIC-III; ``OOD'' refers to training 
on MIMIC-IV and testing on MIMIC-III without fine-tuning. 
``Drop'' shows the relative degradation from In-Domain to OOD. 
\mname{} requires no parameter updates in either regime.}
\label{tab:supervised_full}
\end{table*}

\begin{figure}[t]
    \centering
    \includegraphics[width=0.95\columnwidth]{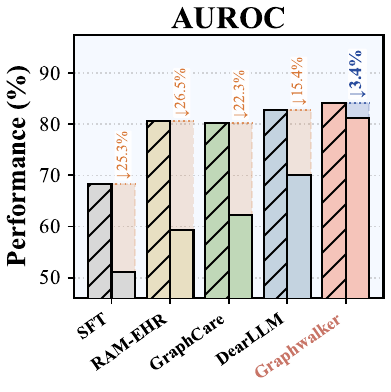}
    \captionsetup{font=footnotesize}
    \caption{In-domain vs.\ cross-dataset OOD AUROC on MIMIC-III 
    in-hospital mortality, complementing 
    Figure~\ref{fig:supervised_comparison} in the main text. 
    Hatched bars show in-domain performance (trained and tested 
    on MIMIC-III); solid bars show OOD performance (trained on 
    MIMIC-IV, tested on MIMIC-III without fine-tuning); shaded 
    regions show the relative drop. \mname{} shows a smaller 
    drop than the supervised baselines we evaluated.}
    \label{fig:supervised_comparison_auroc}
    \vspace{-0.4cm}
\end{figure}
We evaluate all methods on \textbf{in-hospital mortality 
prediction} on MIMIC-III in two regimes:

\paragraph{In-Domain.} 
Each supervised baseline is trained on the MIMIC-III training 
set and evaluated on the MIMIC-III test set, following the 
standard supervised protocol. For \mname{}, the patient graph 
and cohort partitions are constructed on the MIMIC-III 
training set, and predictions are made on the MIMIC-III test 
set without any parameter updates.

\paragraph{Cross-Dataset OOD.} 
Each supervised baseline is trained on the MIMIC-IV training 
set and evaluated on the MIMIC-III test set 
\emph{without any fine-tuning}. For \mname{}, the patient 
graph, cohort partitions, and case retrieval base are all 
constructed on the MIMIC-IV training set, while predictions 
remain on the MIMIC-III test set. The EHR encoder SMART is a 
\textbf{frozen pretrained model} that is not fine-tuned for 
any specific dataset, enabling cross-dataset patient embedding 
comparison without retraining; the frozen LLM backbone and 
all other \mname{} components likewise remain unchanged across 
both regimes.

\paragraph{Baselines.} 
We compare against four supervised paradigms:
\begin{itemize}[leftmargin=*,noitemsep,topsep=2pt]
\item \textbf{SFT} fine-tunes the LLM backbone on the source 
training set with next-token prediction loss to generate the 
binary mortality label.
\item \textbf{RAM-EHR}~\cite{xu2024ram} enhances supervised 
prediction by retrieving knowledge from multiple sources and 
applying consistency regularization.
\item \textbf{GraphCare}~\cite{jiang2024graphcare} acquires 
personalized knowledge by directing LLMs to produce triples, 
which are then fed into a graph neural network predictor.
\item \textbf{DearLLM}~\cite{xu2025dearllm} captures 
quantitative feature correlations through conditional 
perplexity of LLM deduction, augmenting supervised predictors 
via a feature-frequency-aware graph pooling method.
\end{itemize}
All baselines and \mname{} use \textit{Qwen3-14B} as the LLM 
backbone for fair comparison; implementations follow the 
original papers.

\paragraph{Results.} 
Table~\ref{tab:supervised_full} reports detailed in-domain and 
cross-dataset OOD performance, complementing 
Figure~\ref{fig:supervised_comparison} in the main text. We 
summarize three observations:

\textit{(i) In-domain, \mname{} outperforms all four 
supervised baselines on every metric.} For example, against 
the strongest supervised baseline DearLLM, \mname{} attains 
$84.19$ vs.\ $82.77$ AUROC and $59.46$ vs.\ $58.02$ AUPRC, 
despite requiring no parameter updates.

\textit{(ii) Under distribution shift, all supervised methods 
show notable degradation.} The drop ranges from 
$\downarrow$15.4\% to $\downarrow$42.9\% across metrics, with 
SFT degrading the most and DearLLM the least. Notably, even 
hybrid pipelines that leverage LLM-derived knowledge exhibit 
a comparable level of degradation: in these methods, the LLM 
is used as an upstream knowledge source while a downstream 
supervised predictor handles the final prediction, so the 
overall pipeline inherits the generalization limits of its 
supervised component. The reliance on fixed input schemas 
(feature order, availability, and encoding) further renders 
these models brittle to schema variations across institutions 
with heterogeneous EHR systems.

\textit{(iii) \mname{} shows substantially smaller 
degradation, suggesting it learns more transferable clinical 
patterns than feature-level correlations.} The drop ranges 
from $\downarrow$3.4\% (AUROC) to $\downarrow$10.9\% (AUPRC), 
noticeably smaller than that of the supervised baselines we 
evaluated. Beyond the absence of source-specific parameter 
updates, this robustness reflects a more fundamental property 
of our approach: rather than fitting shallow feature-level 
correlations specific to the source dataset, \mname{} 
performs \emph{reasoning by analogy over retrieved patient 
cases}, asking the LLM to interpret each test patient by 
analogy to clinically similar past patients. This case-level 
reasoning relies on transferable clinical patterns---such as 
disease progression and comorbidity dynamics---that 
generalize more naturally across hospitals, populations, and 
EHR schemas than feature-level statistical correlations.

\subsection{RQ4 Results}
\label{app:blackbox_agentic_graphwalker}
\begin{table*}[t!]
\centering
\fontsize{8pt}{9pt}\selectfont
\setlength{\tabcolsep}{5.2pt}
\renewcommand{\arraystretch}{1.05}
\begin{tabular}{l|ccc|ccc}
\toprule
\rowcolor{gray!10}
\textbf{Method}
& \multicolumn{3}{c|}{\textbf{Direct Answer (Few-shot)}}
& \multicolumn{3}{c}{\textbf{+ ReAct}} \\
\rowcolor{gray!10}
& \textbf{AUROC} & \textbf{AUPRC} & \textbf{F1}
& \textbf{AUROC} & \textbf{AUPRC} & \textbf{F1} \\
\midrule
Zero-shot
& \pmstd{70.81}{4.86} & \pmstd{33.69}{6.88} & \pmstd{41.42}{6.20}
& \pmstd{71.11}{5.47} & \pmstd{32.86}{4.97} & \pmstd{43.61}{6.33} \\
Random
& \pmstd{70.41}{4.69} & \pmstd{31.32}{6.94} & \pmstd{39.12}{5.61}
& \pmstd{68.46}{3.26} & \pmstd{29.08}{5.53} & \pmstd{39.08}{4.96} \\
Semantic-emb
& \pmstd{69.91}{5.23} & \pmstd{25.78}{6.32} & \pmstd{36.30}{6.09}
& \pmstd{70.91}{4.11} & \pmstd{26.01}{5.37} & \pmstd{36.92}{5.58} \\
EHR-emb
& \underline{\pmstd{81.59}{4.87}} & \underline{\pmstd{48.45}{6.96}} & \underline{\pmstd{46.07}{5.78}}
& \underline{\pmstd{82.09}{7.81}} & \underline{\pmstd{47.98}{5.96}} & \underline{\pmstd{48.51}{6.45}} \\
\midrule
\textbf{\mname{}}
& \textbf{\pmstd{87.64}{5.14}} & \textbf{\pmstd{62.31}{8.96}} & \textbf{\pmstd{64.79}{7.74}}
& \textbf{\pmstd{88.72}{3.18}} & \textbf{\pmstd{64.69}{2.43}} & \textbf{\pmstd{67.11}{2.45}} \\
\bottomrule
\end{tabular}
\captionsetup{font=footnotesize}
\caption{Results on MIMIC-III Mortality with GPT-5.2 as the 
black-box LLM, comparing two inference settings under the 
\emph{same} few-shot demonstrations: 
\textbf{Direct Answer (Few-shot)} where the LLM directly 
outputs the prediction with no reasoning scaffold, and 
\textbf{+ ReAct} where the LLM applies ReAct-style 
intermediate reasoning before generating the prediction. 
Bold = best within column, underline = second-best. Higher 
is better ($\uparrow$).}
\label{tab:blackbox_react_results}
\vspace{-0.4cm}
\end{table*}
We investigate two practical questions beyond the main 
experiments: (i) \textbf{whether \mname{} can be extended 
to closed-source or black-box LLMs} where token-level 
probabilities are not accessible, and (ii) \textbf{whether 
\mname{} can be combined with agentic reasoning frameworks} 
such as ReAct to yield further gains under the same 
few-shot setting. We implement a black-box variant of 
\mname{} (Sec~\ref{app:blackbox_graphwalker}) and integrate 
it with a ReAct-style scaffold 
(Sec~\ref{app:graphwalker_agentic}).

\subsubsection{Black-box GraphWalker}
\label{app:blackbox_graphwalker}

The original \mname{} estimates the marginal information 
gain of a candidate demonstration via the change in 
conditional entropy computed from model probabilities. 
While effective for open-source LLMs with accessible 
logits, this design is not directly applicable to 
closed-source frontier models exposed only through 
API-based text generation. To address this, we develop a 
black-box variant that replaces entropy-based scoring with 
LLM self-evaluation.

\paragraph{Pipeline.}
The graph construction and search structure remain 
unchanged. We build a patient KNN graph from the training 
set using pretrained EHR embeddings, then apply Leiden 
clustering to obtain patient cohorts. For each test 
patient, we retrieve the top-$L$ most relevant cohorts and 
initialize anchor nodes by selecting the top-$K$ most 
similar patients within each cohort, forming the initial 
frontier for graph-guided demonstration search.

The key difference lies in the greedy selection step. 
Instead of computing $\Delta H$ from token-level 
probabilities, we score each frontier candidate by querying 
the black-box LLM with a self-evaluation prompt. Given the 
target patient, the already selected demonstrations, and 
one candidate demonstration, the model rates how much 
\emph{additional useful information} the candidate provides 
for predicting the target patient's outcome on an integer 
scale from 0 to 10.

\paragraph{Greedy selection with self-evaluation.}
At each iteration, all frontier candidates are scored in 
parallel through API calls. We select the highest-scoring 
candidate, add it to the demonstration set, and expand the 
frontier with its graph neighbors. The process repeats 
until either the maximum number of demonstrations is 
reached or the highest candidate score becomes zero. This 
preserves the cohort-guided frontier expansion and greedy 
search structure of \mname{}, while replacing white-box 
entropy computation with a black-box scoring proxy.

\paragraph{Scoring prompt.}
The scoring prompt 
(Table~\ref{tab:prompt_graphwalker_blackbox_scoring}) 
approximates the same intuition as the original 
information-gain objective: a useful candidate should be 
clinically relevant to the test patient, provide 
complementary evidence beyond the already selected 
examples, and reduce uncertainty about the target 
prediction. Concretely, the black-box scorer assesses each 
candidate from three aspects: (1) clinical similarity or 
complementarity to the target patient, (2) whether it 
contributes information not already covered by selected 
examples, and (3) how much it helps reduce predictive 
uncertainty. The model outputs only a single integer score 
from 0 to 10.

\begin{tcolorbox}[
colback=lightgray!20,
colframe=darkgray!80,
fontupper=\small,
title=Black-box GraphWalker Scoring Prompt
]
\label{tab:prompt_graphwalker_blackbox_scoring}

You are evaluating whether a candidate patient example would be helpful for predicting a test patient's clinical outcome.

\textbf{Task:} \\
\texttt{\{TASK\_DESCRIPTION\}}

\textbf{Already Selected Examples for In-Context Learning:} \\
\texttt{\{SELECTED\_EXAMPLES\}}

\textbf{Candidate Example to Evaluate:} \\
\texttt{\{CANDIDATE\_EXAMPLE\}}

\textbf{Test Patient to Predict:} \\
\texttt{\{TEST\_PATIENT\_DETAIL\}}

\textbf{Instruction:} \\
Rate how much \emph{additional useful information} the candidate example provides for predicting the test patient's outcome, considering:
\begin{itemize}
  \item Clinical similarity or complementarity to the test patient
  \item Whether it provides new information not already covered by selected examples
  \item How much it would help reduce uncertainty about the test patient's outcome
\end{itemize}

\textbf{Score Definition (0--10):}
\begin{itemize}
  \item \texttt{0}: completely redundant or irrelevant
  \item \texttt{5}: moderately helpful, provides some new perspective
  \item \texttt{10}: extremely informative, highly relevant and complementary
\end{itemize}

\textbf{Output Format:} \\
Output \textbf{ONLY} a single integer score from \texttt{0} to \texttt{10}, and nothing else.

\end{tcolorbox}

\paragraph{Discussion.}
This black-box implementation does not claim to exactly 
reproduce the white-box entropy objective. Instead, it 
provides a practical approximation that preserves the core 
design principle of \mname{}: selecting demonstrations 
based on \emph{target-conditioned marginal utility} rather 
than static similarity alone. As a result, \mname{} 
extends to stronger closed-source LLMs without requiring 
access to internal logits, broadening its applicability in 
realistic deployment settings.

\paragraph{Experiments.}
\label{app:blackbox_graphwalker_setting}
We evaluate the black-box variant under a standard 
few-shot setting on \textit{MIMIC-III Mortality}, where 
the downstream LLM directly outputs the final prediction 
without any additional reasoning scaffold.

\paragraph{Results.}
As shown in Table~\ref{tab:blackbox_react_results}, we 
summarize three observations:

\textit{(i) Generic semantic similarity is actively harmful 
under black-box settings.} \textit{Semantic-emb} not only 
substantially underperforms the zero-shot baseline on AUPRC 
and F1 (e.g., AUPRC $25.78$ vs.\ $33.69$), but is also 
worse than \textit{Random} retrieval ($25.78$ vs.\ $31.32$ 
on AUPRC). This indicates that surface-level semantic 
similarity, when divorced from clinical structure, can 
mislead demonstration selection rather than help it.

\textit{(ii) Clinically informed similarity helps but 
remains single-perspective.} \textit{EHR-emb} provides a 
stronger retrieval signal by capturing clinically 
meaningful patient similarity, clearly outperforming both 
zero-shot and random retrieval. Yet as a purely 
similarity-based method, it still leaves substantial 
headroom relative to \mname{}.

\textit{(iii) Black-box \mname{} substantially outperforms 
all single-perspective baselines.} Compared with the 
strongest baseline \textit{EHR-emb}, black-box \mname{} 
improves AUPRC from $48.45$ to $62.31$. This confirms that 
its dual-view advantage, which bridges cohort-level 
clinical similarity and model-side utility estimation, 
generalizes beyond the original entropy-based 
implementation: the core strength lies in cohort-guided, 
target-conditioned demonstration construction rather than 
any specific scoring interface.

\subsubsection{Compatibility with Agentic Reasoning}
\label{app:graphwalker_agentic}
Recent LLM agents show strong reasoning and 
decision-making across many domains~\cite{ferrag2025llm}. 
Rather than positioning \mname{} as an alternative to 
agentic reasoning, we view it as a \textbf{complementary 
upstream plugin}: even powerful agent-style reasoning 
depends on the quality of the provided case context, and 
\mname{}'s cohort-guided, target-conditioned demonstration 
selection can supply more informative examples to better 
unlock downstream agentic reasoning.

\paragraph{Experiments.}
We choose \textit{ReAct}~\cite{yao2022react} as a 
representative agentic reasoning framework: it tightly 
couples intermediate reasoning with final decision making 
and is a widely used backbone for dynamic agent-style 
inference. We keep the few-shot demonstrations selected 
by each method unchanged and only replace the downstream 
inference style from direct answer generation to 
ReAct-style reasoning. This design isolates the effect of 
combining \mname{} with agentic reasoning, directly 
testing whether better case selection can further enhance 
a more dynamic reasoning process under the same few-shot 
setting. Experiments are conducted on \textit{MIMIC-III 
Mortality} using the same black-box LLM and demonstration 
selection baselines as in the previous subsection.

\paragraph{Results.}
As shown in Table~\ref{tab:blackbox_react_results}, we 
summarize three observations:

\textit{(i) \mname{} + ReAct further improves over 
standalone \mname{} across all metrics.} Compared with 
the standard black-box setting, \mname{} + ReAct improves 
AUROC from $87.64$ to $88.72$, AUPRC from $62.31$ to 
$64.69$, and F1 from $64.79$ to $67.11$.

\textit{(ii) ReAct does not consistently help, and 
sometimes hurts, weaker baselines.} For \textit{Random}, 
ReAct \emph{actively degrades} performance across all 
three metrics (e.g., AUROC $70.41 \to 68.46$, AUPRC 
$31.32 \to 29.08$). For \textit{Semantic-emb}, changes 
are marginal. For the stronger \textit{EHR-emb}, gains 
are mixed: AUROC and F1 improve slightly while AUPRC 
drops ($48.45 \to 47.98$).

\textit{(iii) Demonstration quality is a prerequisite for 
agentic reasoning to be effective.} Agentic reasoning 
compounds positively with high-quality demonstrations 
(consistent improvements for \mname{}) but compounds 
neutrally or even negatively with low-quality ones. By 
constructing more informative and clinically coherent 
demonstrations, \mname{} provides a stronger evidence 
base on which agentic reasoning can build, positioning 
it not as an alternative to agentic reasoning but as a 
pluggable upstream module.

\subsection{Applicability to More Tasks}
\label{app:apply2more_task}

To further examine the generality of \mname beyond EHR-centric prediction tasks, we extend our evaluation to three additional medical reasoning benchmarks: \textbf{MedQA}~\cite{jin2021disease}, \textbf{CMB}~\cite{wang2024cmb}, and \textbf{CMB-Clin}~\cite{wang2024cmb}.

\textit{MedQA} consists of medical examination questions designed to assess diagnostic reasoning and clinical decision-making, with each question grounded in a realistic patient case description. Although the full benchmark is multilingual, we focus on the English subset, which contains approximately 12.7K case-based questions and has been widely adopted for evaluating LLM-based medical reasoning.

\textit{CMB} is a large-scale Chinese medical benchmark covering a broad range of clinical specialties and question types. Following~\cite{ding2025promed}, we filter the dataset to retain only case-based questions involving explicit patient scenarios, ensuring its suitability for in-context reasoning and demonstration-based evaluation.

\textit{CMB-Clin}~\cite{wang2024cmb} is the inaugural multi-round medical question-answering benchmark constructed from real-world, complex diagnosis and treatment records. Compared with single-turn medical QA benchmarks, CMB-Clin presents more realistic and longitudinal clinical reasoning challenges, requiring models to handle multi-round interactions, evolving clinical conditions, and more comprehensive diagnostic and treatment decision processes.

Dataset statistics are summarized in Table~\ref{tab:dataset_statistics_medqa_cmb}.

\begin{table}[!ht]
\fontsize{9pt}{9pt}\selectfont
\centering
\setlength{\tabcolsep}{8pt}
\renewcommand{\arraystretch}{1.05}
\begin{tabular}{lccc}
\toprule
\textbf{Split} & \textbf{MedQA} & \textbf{CMB} & \textbf{CMB-Clin} \\
\midrule
Train & 10,178 & 15,465 & 83 \\
Val   & 1,272  & 1,940  & 42 \\
Test  & 1,273  & 1,935  & 83 \\
\midrule
Total & 12,723 & 19,340 & 208 \\
\bottomrule
\end{tabular}
\caption{Dataset statistics for additional medical reasoning benchmarks: MedQA, CMB, and CMB-Clin.}
\label{tab:dataset_statistics_medqa_cmb}
\end{table}

\subsubsection{Experimental Setup}
\label{app:more_task_setup}

In this experimental setting, both \mname and the Semantic-embedding baseline use \textit{Qwen3-Embedding-8B} as the retrieval encoder. For fair comparison, all methods share the same backbone LLM, namely \textit{Qwen3-14B}. 

\begin{table*}[t!]
\centering
\fontsize{9pt}{10.5pt}\selectfont
\setlength{\tabcolsep}{8pt}
\renewcommand{\arraystretch}{1.1}
\begin{tabular}{l|c|c|ccc}
\toprule
\rowcolor{gray!10}
\textbf{Method} 
& \textbf{CMB} 
& \textbf{MedQA} 
& \multicolumn{3}{c}{\textbf{CMB-clin}} \\
\rowcolor{gray!10}
& \textbf{EM} 
& \textbf{EM} 
& \textbf{BLEU-4} 
& \textbf{BLEU-1} 
& \textbf{ROUGE-L} \\
\midrule
Zero-shot 
& \pmstd{80.21}{0.92}
& \pmstd{56.56}{3.59}
& \pmstd{21.06}{1.37}
& \pmstd{59.85}{2.30}
& \pmstd{44.97}{1.73} \\
Random
& \pmstd{81.50}{2.78}
& \pmstd{52.96}{3.53}
& \pmstd{26.85}{1.44}
& \pmstd{66.05}{2.08}
& \pmstd{38.63}{1.77} \\
\midrule
Semantic-emb
& \pmstd{80.52}{2.70}
& \underline{\pmstd{59.58}{3.52}}
& \pmstd{26.86}{1.47}
& \pmstd{64.90}{2.01}
& \pmstd{41.26}{2.07} \\
Time-series
& \pmstd{78.05}{2.90}
& \pmstd{52.11}{3.55}
& \underline{\pmstd{29.05}{10.65}}
& \pmstd{66.56}{2.24}
& \pmstd{40.12}{1.57} \\
\midrule
SPELL
& \pmstd{81.98}{2.73}
& \pmstd{54.10}{3.49}
& \pmstd{28.41}{1.46}
& \pmstd{66.75}{2.10}
& \pmstd{36.82}{1.64} \\
Influence
& \underline{\pmstd{83.50}{2.64}}
& \pmstd{53.02}{3.50}
& \pmstd{23.88}{1.18}
& \pmstd{67.81}{1.08}
& \underline{\pmstd{47.05}{1.57}} \\
IDS
& \pmstd{79.57}{2.81}
& \pmstd{52.59}{3.57}
& \pmstd{22.63}{1.53}
& \pmstd{43.79}{3.08}
& \pmstd{25.58}{1.19} \\
CONE
& \pmstd{80.41}{2.82}
& \pmstd{59.17}{3.51}
& \pmstd{29.04}{1.49}
& \underline{\pmstd{68.10}{1.86}}
& \pmstd{41.10}{1.51} \\
LMS3
& \pmstd{82.02}{2.78}
& \pmstd{54.09}{3.60}
& \pmstd{26.78}{1.51}
& \pmstd{62.58}{2.51}
& \pmstd{38.02}{1.72} \\
Delta-KNN
& \pmstd{83.02}{2.66}
& \pmstd{53.47}{3.42}
& \pmstd{23.09}{1.37}
& \pmstd{63.73}{1.84}
& \pmstd{46.50}{1.79} \\
GradSel
& \pmstd{81.48}{2.75}
& \pmstd{52.93}{3.47}
& \pmstd{24.83}{1.45}
& \pmstd{64.98}{2.08}
& \pmstd{45.13}{1.78} \\
\midrule
\rowcolor[HTML]{F0F6FF}
\textbf{Ours}
& \textbf{\pmstd{84.05}{2.58}}
& \textbf{\pmstd{61.10}{3.46}}
& \textbf{\pmstd{29.68}{1.46}}
& \textbf{\pmstd{71.51}{1.26}}
& \textbf{\pmstd{47.48}{1.38}} \\
\bottomrule
\end{tabular}
\captionsetup{font=footnotesize}
\caption{Performance comparison on CMB, MedQA, and CMB-clin using \textit{Qwen3-14B} as the backbone LLM. CMB and MedQA are evaluated by exact match (EM), while CMB-clin is evaluated by BLEU-4, BLEU-1, and ROUGE-L. Higher values indicate better performance ($\uparrow$). We highlight the \textbf{best} and \underline{second best} results.}
\label{tab:medqa_cmb_results}
\vspace{-0.4cm}
\end{table*}
\subsubsection{Results Analysis}
Overall, the results in Table~\ref{tab:medqa_cmb_results} demonstrate that \mname generalizes well beyond EHR-centric prediction settings and remains consistently effective across diverse medical reasoning tasks. On the two multiple-choice benchmarks, \mname achieves the best EM on both CMB (84.05\%) and MedQA (61.10\%), outperforming all compared baselines, including strong retrieval-based and influence-based selection methods. This indicates that \textbf{\mname is not limited to structured EHR classification, but can also transfer effectively to broader medical reasoning scenarios requiring diagnostic understanding and clinical decision-making.}

More importantly, \mname also shows strong generalization to the open-ended generation benchmark CMB-clin, where it achieves the best performance on all three generation metrics, including BLEU-1, BLEU-4, and ROUGE-L. The gains on CMB-clin suggest that \textbf{the benefits of \mname are not confined to answer selection or label prediction, but also extend to multi-round, free-form medical response generation}, where models must produce more comprehensive and coherent answers grounded in complex clinical contexts.

Taken together, these results verify that \textbf{\mname is a generally applicable in-context demonstration selection framework for medical reasoning, rather than a task-specific solution tailored only to EHR analysis}. Its consistent improvements across both discriminative and generative settings highlight its robustness and suggest that the underlying cohort-guided retrieval and example composition mechanism captures transferable reasoning patterns that remain beneficial across different task formats and evaluation protocols.

\section{Case Study}
\label{app:case_study}

To qualitatively illustrate why \mname{} yields more reliable in-context demonstrations for EHR prediction, we present in Table~\ref{tab:case_study} a representative case from the \textit{MIMIC-III Mortality} test set using \textit{Qwen3-14B} as the backbone LLM, where the \textit{Zero-shot} baseline produces a severely miscalibrated prediction, while \mname{} corrects it.

\section{Experimental Setup of Pilot Study}
\label{appendix:pilot_setup}

To investigate the behavior of different ICL strategies for EHR reasoning, we adopt \textit{Qwen3-14B}~\citep{yang2025qwen3} as the backbone LLM for all pilot experiments. 
We conduct experiments on two widely used, large-scale public EHR benchmarks: \textit{MIMIC-III}~\cite{johnson2016mimic} and \textit{MIMIC-IV}~\cite{johnson2023mimic}, and focus on two representative downstream tasks: \textit{mortality prediction} and \textit{readmission prediction}. 
For each task, the LLM is prompted in a few-shot setting using demonstrations selected by different ICL strategies. 
Detailed dataset statistics, preprocessing procedures, task definitions, and prompt templates are provided in Appendix~\ref{app:dataset} and~\ref{app:prompt}.

\paragraph{Definition of Information Gain $\Delta H$.}
To quantitatively analyze how different demonstration selection strategies affect the model’s understanding of a test query, we measure information gain using an entropy-based criterion.
Unlike the marginal gain formulation used in our method, the $\Delta H$ reported in the pilot study is defined \emph{relative to the zero-shot baseline}.

Formally, let $x_{\text{test}}$ denote an unlabeled test input, and let $\mathcal{S}_k$ denote a demonstration set of size $k$ selected by a given ICL strategy.
We define the conditional entropy of the model’s prediction on $x_{\text{test}}$ under a demonstration composition $\mathcal{S}_k$ as
\begin{equation}
\label{eq:appendix_cond_entropy}
H_\theta(x_{\text{test}} \mid \mathcal{S}_k)
=
H_\theta(x_{\text{test}}, \mathcal{S}_k)
-
H_\theta(\mathcal{S}_k),
\end{equation}
where $H_\theta(x_{\text{test}}, \mathcal{S}_k)$ denotes the cross-entropy of the full prompt (including both the demonstrations and the test query), and $H_\theta(\mathcal{S}_k)$ denotes the cross-entropy of the demonstration composition alone.
This conditional entropy reflects the model’s predictive uncertainty and thus its degree of understanding of the test query under the given context.

We then define the information gain of a demonstration composition $\mathcal{S}_k$ relative to the zero-shot setting as
\begin{equation}
\label{eq:appendix_deltaH}
\Delta H_\theta(\mathcal{S}_k)
=
H_\theta(x_{\text{test}} \mid \emptyset)
-
H_\theta(x_{\text{test}} \mid \mathcal{S}_k),
\end{equation}
where $\emptyset$ denotes the empty demonstration set (i.e., zero-shot prompting).
A larger $\Delta H_\theta(\mathcal{S}_k)$ indicates that the selected demonstrations reduce the model’s uncertainty more effectively compared to the zero-shot baseline.

For each ICL strategy and each $k \in \{1,2,3,4\}$, we compute $\Delta H_\theta(\mathcal{S}_k)$ for every validation instance and report the average information gain over the validation set.

\section{Algorithm}
\label{app:algorithm}
Please see Algorithm~\ref{alg:mname}.
For clarity, Algorithm~\ref{alg:mname} presents the full greedy implementation of \mname{}.
In practice, we adopt a lazy greedy strategy to improve computational efficiency, which is discussed in detail in Appendix~\ref{app:lazy_frontier_greedy}.

\begin{algorithm*}[t]
\centering
\caption{\mname}
\label{alg:mname}
\begin{minipage}{1\linewidth}
\normalsize
\begin{algorithmic}[1]
\Require
Training EHR dataset $\{\mathbf{X}_i\}_{i=1}^N$;
test patient record $\mathbf{X}^{(q)}$;
pretrained EHR encoder $\mathcal{M}_{\text{exp}}$;
LLM $M_\theta$;
graph neighbor size $k_g$;
number of retrieved cohorts $K_c$;
anchors per cohort $K_a$;
demonstration budget $K$.

\vspace{0.5mm}
\State \textcolor{gray}{\# Stage 1: Patient Graph Construction and Cohort Discovery}
\For{$i \gets 1$ to $N$}
    \State $\mathbf{h}_i \gets \mathcal{M}_{\text{exp}}(\mathbf{X}_i)$
\EndFor
\State Construct a (symmetrized) $k_g$-NN patient graph $\mathcal{G}=(\mathcal{V},\mathcal{E})$ over $\{\mathbf{h}_i\}$ using cosine similarity
\State $\{\mathcal{C}_m\}_{m=1}^{M} \gets \textsc{Leiden}(\mathcal{G})$
\For{$m \gets 1$ to $M$}
    \State $\mathbf{z}_m \gets \frac{1}{|\mathcal{C}_m|}\sum_{i\in\mathcal{C}_m}\mathbf{h}_i$
\EndFor

\vspace{0.5mm}
\State \textcolor{gray}{\# Stage 2: Instance-aware Cohort Retrieval and Anchor Initialization}
\State $\mathbf{h}^{(q)} \gets \mathcal{M}_{\text{exp}}(\mathbf{X}^{(q)})$
\State $\mathcal{C}^{(q)}_{\text{ret}} \gets
\operatorname{Top}\text{-}K_c\big(\mathrm{sim}(\mathbf{h}^{(q)}, \mathbf{z}_m)\big)$
\State $\mathcal{F}^{(q)} \gets \emptyset$
\For{each cohort $\mathcal{C}_m \in \mathcal{C}^{(q)}_{\text{ret}}$}
    \State $\mathcal{V}^{(q)}_m \gets
    \operatorname{Top}\text{-}K_a\big(\mathrm{sim}(\mathbf{h}^{(q)}, \mathbf{h}_i)\big),\; v_i \in \mathcal{C}_m$
    \State $\mathcal{F}^{(q)} \gets \mathcal{F}^{(q)} \cup \mathcal{V}^{(q)}_m$
\EndFor

\vspace{0.5mm}
\State \textcolor{gray}{\# Stage 3: LLM-guided Greedy Search with Frontier Expansion}

\State $\mathcal{S}^{(q)} \gets \emptyset$
\State $H_0 \gets H_\theta(\mathbf{X}^{(q)} \mid \mathcal{S}^{(q)})$ 

\While{$|\mathcal{S}^{(q)}| < K$ \textbf{and} $\mathcal{F}^{(q)} \neq \emptyset$}
    \State $v^\ast \gets \texttt{None}$; $\Delta^\ast \gets 0$; $H^\ast \gets H_0$
    \For{each $v_i \in \mathcal{F}^{(q)}$}
        \State $H_i \gets H_\theta(\mathbf{X}^{(q)} \mid \mathcal{S}^{(q)} \cup \{v_i\})$
        \State $\Delta H_\theta(v_i \mid \mathcal{S}^{(q)}) \gets H_0 - H_i$
        \If{$\Delta H_\theta(v_i \mid \mathcal{S}^{(q)}) > \Delta^\ast$}
            \State $v^\ast \gets v_i$; $\Delta^\ast \gets \Delta H_\theta(v_i \mid \mathcal{S}^{(q)})$; $H^\ast \gets H_i$
        \EndIf
    \EndFor
    \If{$v^\ast = \texttt{None}$ \textbf{or} $\Delta^\ast \le 0$}
        \State \textbf{break}
    \EndIf
    \State $\mathcal{S}^{(q)} \gets \mathcal{S}^{(q)} \cup \{v^\ast\}$
    \State $H_0 \gets H^\ast$
    \State $\mathcal{F}^{(q)} \gets \mathcal{F}^{(q)} \setminus \{v^\ast\}$
    \State $\mathcal{F}^{(q)} \gets \mathcal{F}^{(q)} \cup \mathcal{N}(v^\ast)$
    \State $\mathcal{F}^{(q)} \gets \mathcal{F}^{(q)} \setminus \mathcal{S}^{(q)}$
\EndWhile
\State \Return $\mathcal{S}^{(q)}$
\end{algorithmic}
\end{minipage}
\end{algorithm*}

\section{Lazy Greedy Search: Theoretical Justification and Implementation}
\label{app:lazy_frontier_greedy}

Lazy greedy is provably exact when the set objective is monotone 
submodular~\cite{minoux2005accelerated}. 
However, monotone submodularity over the entire power set is a 
strong condition that is difficult to verify in closed form for 
LLM-based objectives. Below we identify a strictly weaker 
condition under which lazy greedy remains exact in our setting, 
discuss approximation guarantees beyond this condition, and 
provide complexity analysis and implementation details.

\paragraph{Sufficient Condition: Chain-wise Diminishing Returns.}
The exactness of lazy greedy as an \emph{algorithm} does not 
require submodularity over the entire power set; it only uses 
cached marginal gains along the specific chain of selected sets 
produced by the search. We formalize this observation as follows:

\begin{tcolorbox}[
    colback=gray!5, colframe=black!50, boxrule=0.5pt, arc=2mm,
    left=2mm, right=2mm, top=1.5mm, bottom=1.5mm,
]
\textbf{Assumption A (Chain-wise Diminishing Returns).} 
For any candidate $v \in \mathcal{V}$ and any chain 
$\mathcal{S}_0 \subseteq \mathcal{S}_1 \subseteq \cdots \subseteq 
\mathcal{S}_K$ of subsets produced by greedy selection, the 
marginal information gain is non-increasing along the chain:
\[
\begin{aligned}
&\Delta_q(v \mid \mathcal{S}_t) \ge \Delta_q(v \mid \mathcal{S}_{t+1}), \\
&\quad \forall t = 0, 1, \ldots, K-1, \; v \notin \mathcal{S}_{t+1}.
\end{aligned}
\]
\end{tcolorbox}

\noindent\textbf{Proposition 1 (Lazy Greedy Exactness).} 
\textit{Under Assumption A, lazy greedy returns the same case 
set as full greedy.}

\begin{proof}
We prove by induction on the iteration $t$ that the case sets 
produced by lazy greedy and full greedy coincide, i.e., 
$\mathcal{S}_t^{\text{lazy}} = \mathcal{S}_t^{\text{full}}$ for 
all $t = 0, 1, \ldots, K$.

\textit{Base ($t = 0$).} Both algorithms start from 
$\mathcal{S}_0 = \emptyset$.

\textit{Inductive step.} Assume $\mathcal{S}_{t-1}^{\text{lazy}} 
= \mathcal{S}_{t-1}^{\text{full}}$, and denote their common value 
as $\mathcal{S}_{t-1}$. At iteration $t$, full greedy selects
\[
v_t^{\text{full}} = \arg\max_{v \in F_{t-1}} \Delta_q(v \mid 
\mathcal{S}_{t-1}),
\]
where $F_{t-1}$ is the current frontier and the marginal gain is 
computed exactly for every candidate.

For lazy greedy, consider any candidate $v$ in the priority queue 
whose cached gain was computed at some earlier iteration 
$t' < t$, i.e., $\Delta_q^{(\text{cache})}(v) = \Delta_q(v \mid 
\mathcal{S}_{t'})$. By the inductive hypothesis, 
$\mathcal{S}_{t'}^{\text{lazy}} = \mathcal{S}_{t'}^{\text{full}} 
\subseteq \mathcal{S}_{t-1}$. Applying Assumption A iteratively 
along the chain $\mathcal{S}_{t'} \subseteq 
\mathcal{S}_{t'+1} \subseteq \cdots \subseteq \mathcal{S}_{t-1}$ 
yields
\[
\Delta_q^{(\text{cache})}(v) = \Delta_q(v \mid \mathcal{S}_{t'}) 
\ge \Delta_q(v \mid \mathcal{S}_{t-1}),
\]
so every cached gain in the priority queue is a valid upper bound 
on the current marginal gain.

Let $v_{\text{top}}$ denote the candidate with the largest cached 
gain. Lazy greedy recomputes its exact gain 
$g_{\text{top}}^{\text{exact}} = \Delta_q(v_{\text{top}} \mid 
\mathcal{S}_{t-1})$ and accepts it as $v_t^{\text{lazy}}$ when 
$g_{\text{top}}^{\text{exact}}$ is no smaller than the next cached 
value $g_{\text{next}}^{\text{cache}}$ in the queue. By the upper 
bound property, for every other candidate $v' \in F_{t-1} 
\setminus \{v_{\text{top}}\}$,
\[
g_{\text{next}}^{\text{cache}} \ge 
\Delta_q^{(\text{cache})}(v') \ge 
\Delta_q(v' \mid \mathcal{S}_{t-1}).
\]
Combining these inequalities,
\[
g_{\text{top}}^{\text{exact}} \ge g_{\text{next}}^{\text{cache}} 
\ge \Delta_q(v' \mid \mathcal{S}_{t-1}), \quad \forall v' \in 
F_{t-1} \setminus \{v_{\text{top}}\},
\]
so $v_{\text{top}}$ achieves the maximum exact marginal gain 
over $F_{t-1}$. Therefore $v_t^{\text{lazy}} = v_t^{\text{full}}$, 
and $\mathcal{S}_t^{\text{lazy}} = \mathcal{S}_t^{\text{full}}$.
\end{proof}

Assumption A is strictly weaker than monotone submodularity over 
the entire power set: it requires diminishing returns only along 
the chain of subsets actually encountered by the algorithm, not 
for all subset pairs. Our analysis in 
Section~\ref{sec:pilot_study} (Figure~\ref{fig:entropy}) is 
consistent with Assumption A: marginal gains exhibit clear 
diminishing returns as more cases are added, with 
information-gain-based methods saturating monotonically and 
semantic-similarity-based methods decaying due to redundancy 
among clinically similar cases.

\paragraph{Approximation Guarantee Beyond Assumption A.}
For settings where Assumption A may be violated, the broader 
theory of approximate submodular maximization provides formal 
guarantees. The 
submodularity ratio of $f_q$ is defined as
\[
\gamma_q = \min_{\mathcal{S} \subseteq \mathcal{V}, \, v \notin 
\mathcal{S}} \frac{\Delta_q(v \mid \mathcal{S})}{\Delta_q(v \mid 
\emptyset)},
\]
where $\gamma_q = 1$ recovers strict submodularity. Greedy 
selection achieves a $(1 - e^{-\gamma_q})$ approximation of the 
optimal case set, which degrades gracefully as $\gamma_q$ 
deviates from $1$. This provides a formal fallback guarantee for 
the approximate-submodular regime, ensuring that the quality of 
the lazy greedy solution remains bounded even when Assumption A 
holds only approximately.

\paragraph{Motivation and Complexity.}
A naive implementation of Algorithm~\ref{alg:mname} performs a 
\emph{full greedy} search, recomputing the marginal information 
gain $\Delta_q(\cdot \mid \mathcal{S}^{(q)}_{t-1})$ for every 
candidate in the frontier at each iteration. Let $|F^{(q)}_t|$ 
denote the frontier size at step $t$ and $K$ be the case budget. 
Full greedy therefore requires $\sum_{t=1}^{K} |F^{(q)}_t|$ exact 
evaluations, which translates to $O(K\bar{F})$ LLM forward passes, 
where $\bar{F}$ denotes the average frontier size. Since each 
exact evaluation requires a forward pass over the in-context 
prompt, this cost becomes prohibitive for large patient graphs.

Under Proposition 1 (or approximate submodularity), lazy greedy 
reduces this to approximately $O(|F^{(q)}_0| + R)$ exact 
evaluations, where $R$ is the number of cached entries that 
require re-evaluation during the search. The priority-queue 
overhead is $O(\log |F^{(q)}_t|)$ per update, which is negligible 
compared to LLM inference.

\paragraph{Implementation Details.}
Our implementation follows the standard lazy greedy template 
with a max-priority queue. At each step, we pop the candidate 
with the largest \emph{cached} marginal gain (an upper bound 
under Proposition 1), recompute its \emph{exact} marginal gain 
with respect to the current case set $\mathcal{S}^{(q)}$, and 
compare it with the next cached best. If it remains the best, 
we accept it; otherwise, we update its cached key and push it 
back. Upon selecting $v^\ast$, we update $\mathcal{S}^{(q)} 
\leftarrow \mathcal{S}^{(q)} \cup \{v^\ast\}$, remove $v^\ast$ 
from the frontier, and expand the frontier by adding 
$\mathcal{N}(v^\ast) \setminus \mathcal{S}^{(q)}$. We further 
apply early stopping when the best marginal gain becomes 
non-positive, i.e., $\Delta_q(v^\ast \mid \mathcal{S}^{(q)}) 
\le 0$, preventing the inclusion of misleading or uninformative 
cases.

\section{Prompt Details}
\label{app:prompt}
In this section, we detail the prompt designs used in our framework.
We implement two prompting schemes: a \textit{zero-shot} template, which provides no in-context examples and requires the LLM to directly produce the task-specific output, and a \textit{few-shot} template, which augments the prompt with a set of labeled patient demonstrations to support ICL.
Both templates share a unified structure for presenting longitudinal EHR data.
In addition, we specify task-specific instructions for each prediction task, including in-hospital mortality, readmission, and length-of-stay prediction, ensuring consistent and fair evaluation across different prompting settings.

\begin{tcolorbox}
[colback=lightgray!20,
 colframe=darkgray!80,
 fontupper=\small,
 title=Prompt for Mortality Prediction]
\label{tab:prompt_mortality}

You are tasked with predicting in-hospital mortality based on patient EHR data.

Provide only a floating-point number between 0 and 1 representing the predicted probability of mortality (a higher value indicates a higher likelihood of death).

Do \textbf{not} provide any reasoning, explanation, or additional text. Output \textbf{only} the numerical value.

\textbf{Example:} \texttt{0.XX}

\end{tcolorbox}

\begin{tcolorbox}
[colback=lightgray!20,
 colframe=darkgray!80,
 fontupper=\small,
 title=Prompt for Readmission Prediction]
\label{tab:prompt_readmission}

You are tasked with predicting whether a patient will be readmitted within 30 days after hospital discharge based on EHR data.

Provide only a floating-point number between 0 and 1 representing the predicted probability of 30-day readmission after discharge (including cases where the patient dies within 30 days, which are counted as readmission events).

Do \textbf{not} provide any reasoning, explanation, or additional text. Output \textbf{only} the numerical value.

\textbf{Example:} \texttt{0.XX}

\end{tcolorbox}

\begin{tcolorbox}
[colback=lightgray!20,
 colframe=darkgray!80,
 fontupper=\small,
 title=Prompt for Length-of-Stay (LOS) Prediction]
\label{tab:prompt_los}

You are tasked with predicting the patient’s length of hospital stay based on EHR data.

Provide only a single letter (\textbf{A}, \textbf{B}, \textbf{C}, or \textbf{D}) representing the predicted length-of-stay category:
\begin{itemize}
  \item \textbf{A}: Less than 3 days ($<3$ days)
  \item \textbf{B}: 3 to 7 days ($3$--$7$ days)
  \item \textbf{C}: 7 to 14 days ($7$--$14$ days)
  \item \textbf{D}: More than 14 days ($>14$ days)
\end{itemize}

Do \textbf{not} provide any reasoning, explanation, or additional text. Output \textbf{only} the letter (\textbf{A}, \textbf{B}, \textbf{C}, or \textbf{D}).

\textbf{Example:} \texttt{B}

\end{tcolorbox}

\newpage
\begin{tcolorbox}
[colback=lightgray!20,
 colframe=darkgray!80,
 fontupper=\small,
 title=Zero-shot Prompt Template]
\label{tab:prompt_zero_shot}

You will be provided with longitudinal electronic health record (EHR) data for a single patient.
Each clinical feature is represented as a time-ordered list of measurements corresponding to the same hospital stay.
Missing values are denoted as \texttt{NaN}. Units and reference ranges are provided where applicable.

\textbf{Patient Information:}
\begin{itemize}
  \item Number of measurements: \texttt{\{LENGTH\}}
  \item Measurement times (hours since admission): \texttt{[\{RECORD\_TIME\_LIST\}]}
\end{itemize}

\textbf{Task Description:} \\
\texttt{\{TASK\_DESCRIPTION\}}

\textbf{Clinical Features Over Time:}
\begin{itemize}
  \item \textbf{Heart Rate} (Unit: bpm. Reference range: 60--100): \\
  \texttt{[82, 86, 90, 95, NaN, 88, 84, \dots]}
  
  \item \textbf{Systolic Blood Pressure} (Unit: mmHg. Reference range: $<120$): \\
  \texttt{[110, 118, 125, 132, 120, NaN, 115, \dots]}
  
  \item \textbf{Oxygen Saturation} (Unit: \%. Reference range: 95--100): \\
  \texttt{[100, 99, 98, NaN, 96, 95, 97, \dots]}
  
  \item \textbf{Glasgow Coma Scale (Total)} (Unit: /. Reference range: /.): \\
  \texttt{[15, 14, NaN, 12, 10, 8, \dots]}
  
  \item \textbf{$\vdots$}
\end{itemize}
\textbf{Your Answer:} \\
\texttt{\{MODEL\_OUTPUT\}}
\end{tcolorbox}

\begin{tcolorbox}
[colback=lightgray!20,
 colframe=darkgray!80,
 fontupper=\small,
 title=Few-shot Prompt Template]
\label{tab:prompt_few_shot}

You will be provided with longitudinal electronic health record (EHR) data for a single patient.
Each clinical feature is represented as a time-ordered list of measurements corresponding to the same hospital stay.
Missing values are denoted as \texttt{NaN}. Units and reference ranges are provided where applicable.

\textbf{Patient Information:}
\begin{itemize}
  \item Number of measurements: \texttt{\{LENGTH\}}
  \item Measurement times (hours since admission): \texttt{[\{RECORD\_TIME\_LIST\}]}
\end{itemize}

\textbf{Task Description:} \\
\texttt{\{TASK\_DESCRIPTION\}}

\textbf{Instructions \& Output Format:} \\
\texttt{\{RESPONSE\_FORMAT\}}

\textbf{In-context Examples:} \\
Below are example patient records and their corresponding labels. These examples are provided to guide the prediction for the target patient.

\begin{itemize}
  \item \textbf{Example 1:} \\
  Clinical Features: \texttt{[\dots]} \\
  Label: \texttt{\{LABEL\_1\}}

  \item \textbf{Example 2:} \\
  Clinical Features: \texttt{[\dots]} \\
  Label: \texttt{\{LABEL\_2\}}

  \item \textbf{$\vdots$}
\end{itemize}

\textbf{Target Patient:}

\textbf{Clinical Features Over Time:}
\begin{itemize}
  \item \textbf{Heart Rate} (Unit: bpm. Reference range: 60--100): \\
  \texttt{[82, 86, 90, 95, NaN, 88, 84, \dots]}
  
  \item \textbf{Systolic Blood Pressure} (Unit: mmHg. Reference range: $<120$): \\
  \texttt{[110, 118, 125, 132, 120, NaN, 115, \dots]}
  
  \item \textbf{Oxygen Saturation} (Unit: \%. Reference range: 95--100): \\
  \texttt{[100, 99, 98, NaN, 96, 95, 97, \dots]}
  
  \item \textbf{Glasgow Coma Scale (Total)} (Unit: /. Reference range: /.): \\
  \texttt{[15, 14, NaN, 12, 10, 8, \dots]}
  
  \item \textbf{$\vdots$}
\end{itemize}

\textbf{Your Answer:} \\
\texttt{\{MODEL\_OUTPUT\}}

\end{tcolorbox}

\newcommand{\badpred}[1]{\fcolorbox{red!60!black}{red!10}{\strut #1}}
\newcommand{\goodpred}[1]{\fcolorbox{green!50!black}{green!10}{\strut #1}}

\tcbset{
  gwexample/.style={
    breakable,
    colback=white,
    colframe=black!10,
    boxrule=0.4pt,
    arc=1pt,
    left=6pt,right=6pt,top=4pt,bottom=4pt,
    boxsep=2pt
  }
}

\clearpage
\onecolumn
\renewcommand{\arraystretch}{1.12}
\setlength{\tabcolsep}{6pt}

\begin{center}
\small
\begin{longtable}{
  >{\raggedright\arraybackslash}p{2.4cm}
  >{\raggedright\arraybackslash}p{13.8cm}
}
\caption{Case study on \textit{MIMIC-III Mortality}.}
\label{tab:case_study}\\
\toprule
\rowcolor{gray!10}
\textbf{Component} & \textbf{Content} \\
\midrule
\endfirsthead

\toprule
\rowcolor{gray!10}
\textbf{Component} & \textbf{Content} \\
\midrule
\endhead

\bottomrule
\endfoot

\bottomrule
\endlastfoot

\textbf{Prompt} &
You will be provided with longitudinal electronic health record (EHR) data for a single patient.
Each clinical feature is represented as a time-ordered list of measurements corresponding to the same hospital stay.
Missing values are denoted as \texttt{NaN}.

\vspace{5pt}
\textbf{Patient Information:}
\begin{itemize}\setlength{\itemsep}{-2pt}\setlength{\topsep}{2pt}
  \item Number of measurements: \texttt{19}
  \item Measurement times (hours since admission): \texttt{[4.68, 5.18, 6.18, 7.18, 8.18, \dots]}
\end{itemize}

\vspace{5pt}
\textbf{Task Description:}
You are tasked with predicting in-hospital mortality based on patient EHR data.

\vspace{5pt}
\textbf{Instructions \& Output Format:}
Provide only a floating-point number between 0 and 1 representing the predicted probability of mortality.
Do \textbf{not} provide any reasoning or additional text. Output \textbf{only} the numerical value (e.g., \texttt{0.XX}). \\
\midrule

\textbf{Target Patient} &
\textbf{Clinical Features Over Time:}
\begin{itemize}\setlength{\itemsep}{0pt}\setlength{\topsep}{2pt}
  \item \textbf{Heart Rate (bpm):} \texttt{[117.0, 110.0, 111.0, 107.0, 102.0, \dots, 119.0, 122.0, 118.0]}
  \item \textbf{Systolic BP (mmHg):} \texttt{[99.0, 96.0, 92.0, 109.0, 87.0, \dots, 127.0, 118.0, 120.0]}
  \item \textbf{Diastolic BP (mmHg):} \texttt{[39.0, 56.0, 44.0, 54.0, 47.0, \dots, 87.0, 75.0, 67.0]}
  \item \textbf{Mean BP (mmHg):} \texttt{[59.0, 69.3, 60.0, 72.3, 60.3, \dots, 100.3, 89.3, 84.7]}
  \item \textbf{SpO\textsubscript{2} (\%):} \texttt{[97.0, 94.0, 95.0, 95.0, 94.0, \dots, 96.0, 95.0, 95.0]}
  \item \textbf{GCS (Total):} \texttt{[NaN, 15.0, 15.0, 15.0, 15.0, \dots, 15.0, 15.0, 15.0]}
  \item[]\vspace{-6pt}\textbf{$\vdots$}
\end{itemize} \\
\midrule

\textbf{Ground Truth} & \texttt{0} (Survival). \\
\midrule

\textbf{Zero-shot Output} &
\textbf{Prediction:} \texttt{0.72} \quad $\rightarrow$ \badpred{\textbf{Incorrect}} \\
\midrule

\textbf{\mname{} Examples} &
\textbf{Example 1} \hfill \textit{Label: \texttt{0}; }
\begin{itemize}\setlength{\itemsep}{1pt}\setlength{\topsep}{0pt}
  \item \textbf{Heart Rate:} \texttt{[78.0, 104.0, 81.0, 98.0, 74.0, \dots, 73.0, NaN, 74.0, 72.0]}
  \item \textbf{Systolic BP:} \texttt{[119.0, 134.0, 111.0, 165.0, 149.0, \dots, 107.0, 96.0, 112.0, 118.0]}
  \item \textbf{Diastolic BP:} \texttt{[62.0, 70.0, 79.0, 73.0, 66.0, \dots, 43.0, 56.0, 62.0]}
  \item \textbf{SpO\textsubscript{2}:} \texttt{[98.0, 94.0, 95.0, 98.0, 96.0, \dots, 98.0, NaN, 97.0, 96.0]}
  \item \textbf{GCS (Total):} \texttt{[NaN, 14.0, NaN, 14.0, NaN, \dots, 15.0, \dots]}
  \item[]\vspace{-6pt}\textbf{$\vdots$}\vspace{-4pt}
\end{itemize}

\vspace{2pt}\hrule\vspace{4pt}

\textbf{Example 2} \hfill \textit{Label: \texttt{0};  }
\begin{itemize}\setlength{\itemsep}{1pt}\setlength{\topsep}{0pt}
  \item \textbf{Heart Rate:} \texttt{[87.0, 92.0, 92.0, 97.0, 90.0, \dots, 88.0, 77.0, 75.0, 74.0]}
  \item \textbf{Systolic BP:} \texttt{[138.0, 155.0, 143.0, 152.0, 131.0, \dots, 136.0, 116.0, 133.0, 136.0]}
  \item \textbf{Diastolic BP:} \texttt{[62.0, 78.0, 69.0, 79.0, 57.0, \dots, 59.0, 66.0, 61.0]}
  \item \textbf{SpO\textsubscript{2}:} \texttt{[93.0, 93.0, 93.0, 95.0, 91.0, \dots, 93.0, 97.0, 98.0, 98.0]}
  \item \textbf{GCS (Total):} \texttt{[NaN, 15.0, NaN, \dots, 15.0, \dots]}
  \item[]\vspace{-6pt}\textbf{$\vdots$}\vspace{-4pt}
\end{itemize}

\vspace{2pt}\hrule\vspace{4pt}

\textbf{Example 3} \hfill \textit{Label: \texttt{0};  }
\begin{itemize}\setlength{\itemsep}{1pt}\setlength{\topsep}{0pt}
  \item \textbf{Heart Rate:} \texttt{[102.0, 105.0, 95.0, 91.0, 96.0, \dots, 77.0, 79.0, 80.0, 81.0]}
  \item \textbf{Systolic BP:} \texttt{[122.0, 118.0, 115.0, 118.0, 106.0, \dots, 132.0, 118.0, 129.0, 130.0]}
  \item \textbf{Diastolic BP:} \texttt{[62.0, 52.0, 65.0, 74.0, 63.0, \dots, 84.0, 76.0, 79.0, 85.0]}
  \item \textbf{SpO\textsubscript{2}:} \texttt{[83.0, 95.0, 94.0, 95.0, 93.0, \dots, 93.0, 93.0, 92.0, 91.0]}
  \item \textbf{GCS (Total):} \texttt{[NaN, NaN, 15.0, \dots, 15.0, \dots]}
  \item[]\vspace{-6pt}\textbf{$\vdots$}
\end{itemize}
\\
\midrule

\textbf{\mname{} Output} &
\textbf{Prediction:} \texttt{0.0689} \quad $\rightarrow$ \goodpred{\textbf{Correct}}\quad 
\end{longtable}
\end{center}

\clearpage
\twocolumn
\renewcommand{\arraystretch}{1.0}

\section{Notations Table}
\label{sec:app_notation}
This section summarizes the key notations used in the \mname framework.

\begin{table}[htbp]
\centering
\small
\begin{tabular}{p{2.0cm} p{5.2cm}}
\toprule
\rowcolor{gray!10}
\textbf{Symbol} & \textbf{Description} \\
\midrule

$\mathbf{X}_i$ 
& EHR record of patient $i$ \\

$\mathcal{M}_{\mathrm{exp}}$ 
& Pretrained EHR encoder \\

$\mathbf{h}_i$ 
& Patient embedding of patient $i$ \\

$N$ 
& Number of patients in the EHR base \\

\hline

$\mathcal{G}$ 
& Population-level patient graph \\

$\mathcal{V}$ 
& Set of patient nodes in the graph \\

$\mathcal{E}$ 
& Set of edges constructed via $k_g$-nearest neighbors in the embedding space \\

$k_g$ 
& Number of nearest neighbors used for graph construction \\

$\mathrm{sim}(\cdot,\cdot)$ 
& Cosine similarity function between embeddings \\

\hline

$\mathcal{C}_m$ 
& Cohort subgraph discovered from the patient graph via Leiden algorithm \\

$M$ 
& Total number of discovered cohorts \\

$\mathbf{z}_m$ 
& Centroid embedding of cohort $\mathcal{C}_m$ obtained by mean pooling \\

$\mathcal{Q}$ 
& Graph modularity objective optimized during cohort discovery \\

\hline

$\mathbf{X}^{(q)}$ 
& EHR record of the target patient \\

$\mathbf{h}^{(q)}$ 
& Embedding of the target patient \\

$\mathcal{C}^{(q)}_{\mathrm{ret}}$ 
& Set of top-$K_c$ cohorts retrieved for the target patient\\

$K_c$ 
& Number of cohorts retrieved for the target patient \\

\hline

$\mathcal{V}^{(q)}_m$ 
& Set of anchor nodes selected from cohort $\mathcal{C}_m$ for the target patient \\

$K_a$ 
& Number of anchor nodes selected per retrieved cohort \\

$\mathcal{F}^{(q)}$ 
& Search frontier consisting of candidate demonstration nodes for the target patient \\

\hline

$\mathcal{S}^{(q)}$ 
& Current demonstration set constructed for the target patient \\

$K$ 
& Maximum number of demonstrations \\

$v_i$ 
& Candidate demonstration node corresponding to patient $i$ \\

$v^\ast$ 
& Selected demonstration node that maximizes marginal information gain \\

\hline

$x_{\mathrm{test}}$ 
& Unlabeled test query constructed from the target patient’s EHR record \\

$H_\theta(x_{\mathrm{test}} \mid \mathcal{S})$
& Conditional entropy of the LLM’s prediction on the test query given demonstration set $\mathcal{S}$ \\

$\Delta H_\theta(v_i \mid \mathcal{S})$ 
& Marginal information gain obtained by adding candidate $v_i$ to demonstration set $\mathcal{S}$ \\

$\theta$ 
& Parameters of the target LLM \\

\hline

$\mathcal{N}(v)$ 
& Neighbor set of node $v$ in the patient graph, used for frontier expansion \\

\bottomrule
\end{tabular}
\caption{Key Notations Used in \mname}
\label{tab:notations}
\end{table}

\section{Code and Data Availability}
\label{sec:code_data_availability}
To support reproducibility and facilitate future research, we will publicly release the full implementation of \mname{} along with the processed datasets used in our experiments upon publication.

\section{Computational Resources and Software Environment}
\label{app:compute}

All experiments were conducted on a server equipped with two NVIDIA H20 GPUs (96~GB memory each) and 503~GB system RAM, running Ubuntu~22.04.5~LTS.
The software environment was based on Python~3.11.11 with Conda~23.5.2, and experiments were implemented using PyTorch~2.6.0, HuggingFace Transformers~4.51.3, and SpaCy~3.8.4, all with default settings unless otherwise specified.

\section{Use of Large Language Models}
\label{sec:use_of_llms}
In this work, large language models (LLMs) were used in a supportive role to assist with language refinement and programming-related tasks, including improving clarity, grammatical correctness, and presentation quality, as well as providing high-level guidance for code structuring and debugging.
All LLM-assisted outputs were carefully reviewed, verified, and, when necessary, revised by the authors prior to inclusion.
The core research ideas, methodological design, experimental setup, and result analysis were conceived and carried out entirely by the authors.
LLMs did not contribute to the formulation of scientific hypotheses, the design of the proposed methods, or the derivation of research conclusions.
\end{document}